%% file: main.tex
\documentclass[sigconf]{acmart}

\AtBeginDocument{
  }

\usepackage{multirow}
\usepackage{colortbl}
\usepackage{amsmath}
\usepackage{algorithm}
\usepackage{algpseudocode}
\usepackage{titletoc}

\usepackage{xcolor}
\algrenewcommand\algorithmiccomment[1]{\hfill{\color[gray]{0.3}$\triangleright$ #1}}
\usepackage{pifont}

\newcommand{\mainsecref}[1]{Section~\ref{#1}}
\newcommand{\mainequref}[1]{Eq.~\eqref{#1}}
\newcommand{\mainalgref}[1]{Algorithm~\ref{#1}}
\newcommand{\mainequrefrange}[2]{Eq.~\eqref{#1}--\eqref{#2}}
\usepackage{tcolorbox}
\usepackage{booktabs}

\copyrightyear{2026}
\acmYear{2026}
\setcopyright{cc}
\setcctype{by}
\acmConference[MM '26]{Proceedings of the 34th ACM International Conference on Multimedia}{November 10--14, 2026}{Rio de Janeiro, Brazil}
\acmBooktitle{Proceedings of the 34th ACM International Conference on Multimedia (MM '26), November 10--14, 2026, Rio de Janeiro, Brazil}
\acmDOI{10.1145/3767308.3836033}
\acmISBN{979-8-4007-2213-4/2026/11}

\begin{document}

\title{STAR-VF: Spatio-Temporal Aware Representation Condensation for Data-Efficient Video Fusion}

\author{Yuchen Xian}
\orcid{0009-0005-0254-8365}
\affiliation{
  \institution{Zhejiang University}
  \city{Hangzhou}
  \state{Zhejiang}
  \country{China}
}
\affiliation{
  \institution{Nanyang Technological University}
  \country{Singapore}
}
\email{yuchen_xian@zju.edu.cn}

\author{Yunqiu Xu}
\authornote{Corresponding author.}
\orcid{0000-0002-2940-4805}
\affiliation{
  \institution{National University of Singapore}
  \country{Singapore}}
\email{imyunqiuxu@gmail.com}

\author{Guosheng Lin}
\orcid{0000-0002-0329-7458}
\affiliation{
  \institution{Nanyang Technological University}
  \country{Singapore}
}
\email{gslin@ntu.edu.sg}

\author{Yang He}
\orcid{0000-0002-2257-6073}
\affiliation{
 \institution{CFAR, Agency for Science, Technology and Research}
 \country{Singapore}
}
\email{he_yang@a-star.edu.sg}

\renewcommand{\shortauthors}{Yuchen Xian, Yunqiu Xu, Guosheng Lin, and Yang He}

\input{section/0_abstract}




\maketitle

\input{section/1_introduction}
\input{section/2_relatedwork}
\input{section/3_preliminary}

\input{section/4_methodology}

\input{section/5_experiments}

\input{section/6_conclusion}

\begin{acks}
This research is supported in part by the RIE2025 Industry Alignment Fund – Industry Collaboration Projects (IAF-ICP)
(Award I2301E0026), administered by A*STAR, as well as supported by Alibaba Group and NTU Singapore through
Alibaba-NTU Global e-Sustainability CorpLab (ANGEL).
This research is also supported in part by the A*STAR Career Development Fund (CDF) under Grant C243512011.
\end{acks}

\bibliographystyle{ACM-Reference-Format}
\balance
\bibliography{sample-base}

\clearpage
\onecolumn
\appendix
\input{section/x_appendix}

\end{document}

%% file: section/0_abstract.tex
\begin{abstract}
Video fusion aims to integrate complementary cues from heterogeneous sources, producing visually informative videos with coherent spatio-temporal structures.
Jointly learning cross-modal complementarity and temporal dependencies typically relies on large-scale, high-quality video datasets, whose storage and repeated use during optimization incur substantial costs, making data-efficient learning highly desirable.
To address this challenge, we rethink video fusion from the perspective of dataset condensation and propose STAR-VF, the first framework that condenses spatio-temporal representations for data-efficient video fusion. 
STAR-VF condenses the original training set into a compact set of learnable spatio-temporal representations, which are decoded into synthetic video clips and optimized using feature-level supervision provided by a frozen proxy model.
The resulting condensed data preserves fusion-relevant spatial and temporal cues without requiring ground-truth fused videos as supervision. To adapt fusion learning to condensed representations, we equip the final fusion model with flow-free inter-frame temporal modeling and query-driven intra-frame aggregation, enabling direct temporal
interaction and adaptive selection of fusion-relevant information in representation space.
Extensive experiments on infrared-visible, multi-exposure, multi-focus, and medical video fusion benchmarks demonstrate that STAR-VF, trained with only approximately 1/16 of the original temporal training set, achieves fusion quality comparable to methods trained on the full data.
Further analyses validate the contributions of learned condensation and flow-free temporal modeling, and demonstrate the robustness of the condensed supervision across different proxy encoders and downstream fusion models.
Project page: \url{https://zju-xyc.github.io/STAR-VF-Project-Page/}.
\end{abstract}

%% file: section/1_introduction.tex
\section{Introduction}

Video fusion~\cite{zhao2025unified, tang2025videofusion,gong2025temcoco,zhao2026mambavf,11644855,xie2024rcvs,li2026drfusion} is an important yet under-explored task that aims to integrate video sequences from different sensors or vary acquisition conditions into a unified and temporally consistent output, yielding a more informative representation for dynamic scenes.
Compared to static image fusion~\cite{xie2024rcvs,tang2018pixel,xu2020u2fusion,ma2019fusiongan,ma2022swinfusion,yi2024diff,WU2023116982,11267268,11078301,liu2025dcevo,xian20262dgrids1dtokens,li2026unifusion}, video fusion is more challenging, as it must jointly integrate complementary heterogeneous information at the frame level while preserving content stability and temporal structural consistency across frames.

Despite recent advances in video fusion, prior works~\cite{zhao2025unified,tang2025videofusion,gong2025temcoco,zhao2026mambavf} still follow a full-data training paradigm, learning spatial complementarity and temporal dependency from complete aligned video pairs. 
While effective, this paradigm implicitly assumes continuous access to large-scale raw spatiotemporal training data.
For video fusion, however, the challenge lies not only in the larger scale and higher training cost of videos, but also in the dual nature of the supervision required for effective fusion: the model must capture complementary information across heterogeneous input sources while preserving temporal coherence over time \cite{zhao2025unified,gong2025temcoco,li2024videomamba}. Once the training data is compressed, both signals can be degraded simultaneously, weakening the model's ability to learn reliable fusion patterns~\citep{li2025leveraging}. Therefore, the central question for dataset compression in video fusion is not whether less data can be used, but \textbf{whether the spatio-temporal supervision essential to effective fusion can still be preserved} after substantial data reduction. 

\input{Figure/temporal}

Therefore, \textbf{applying dataset condensation to video fusion introduces three key challenges}. \ding{182} Conventional temporal alignment based on optical flow estimation~\cite{teed2020raftrecurrentallpairsfield,basicvsr} and feature warping~\cite{RevisitingTemporal} is designed for raw frames or dense features, making it poorly matched to compact condensed representations. \ding{183} In the condensed setting, fusion-critical cues are no longer expected to be uniformly distributed over all elements, so uniform aggregation can easily dilute the most informative components. \ding{184} After substantial data reduction, the condensed training source must still preserve the fusion-relevant supervision required for both cross-source complementarity and temporal coherence. However, these challenges remains largely underexplored. Existing dataset condensation methods are primarily designed for static tasks such as image classification~\cite{ding2024condtsf,UniDD}. Therefore, making dataset condensation truly applicable to video fusion requires not only compressing the training data, but also developing a fusion framework that preserves spatio-temporal supervision and remains compatible with condensed representations.

To address this issue, we propose \textbf{STAR-VF}, a \textbf{S}patio-\textbf{T}emporal \textbf{A}ware \textbf{R}epresentation condensation framework for data-efficient \textbf{V}ideo \textbf{F}usion. STAR-VF compresses the original training set into a small set of learnable compact spatio-temporal representations, which are decoded into synthetic video clips under feature-level supervision from a frozen proxy network to preserve fusion-relevant signals. Built on this condensed set, the final fusion model is further designed to operate effectively on compact representations, with \emph{flow-free inter-frame temporal modeling} for direct temporal dependency learning in compact space and \emph{query-driven intra-frame aggregation} for adaptive extraction of fusion-critical intra-frame information, shown in Figure~\ref{fig:temporal_alignment}. Extensive experiments on four representative video fusion tasks show that, using only condensed synthetic data amounting to 1/16 of the original dataset, STAR-VF still achieves performance close to full-data training. These results suggest that spatio-temporal representation condensation can serve as a viable data-efficient learning paradigm for video fusion. The main contributions of this paper are as follows:

% \vspace{-4pt}
\begin{enumerate}
    \item We first revisit video fusion from a data-efficient learning perspective and approach it through the lens of dataset condensation, aiming to enable efficient training while retaining fusion-critical spatio-temporal information.
    \item We propose STAR-VF, a new video fusion framework with spatio-temporal aware representation compression, which learns a condensed training source and adapts fusion to condensed representations via flow-free inter-frame temporal modeling and query-driven intra-frame aggregation.
    \item We evaluate STAR-VF on four video fusion tasks, validating that training on condensed data ($1/16$ of the original dataset) still achieves performance competitive with full-data-trained state-of-the-art approaches.
\end{enumerate}

%% file: Figure/temporal.tex
\begin{figure}[t]
    \centering
    \includegraphics[width=\columnwidth]{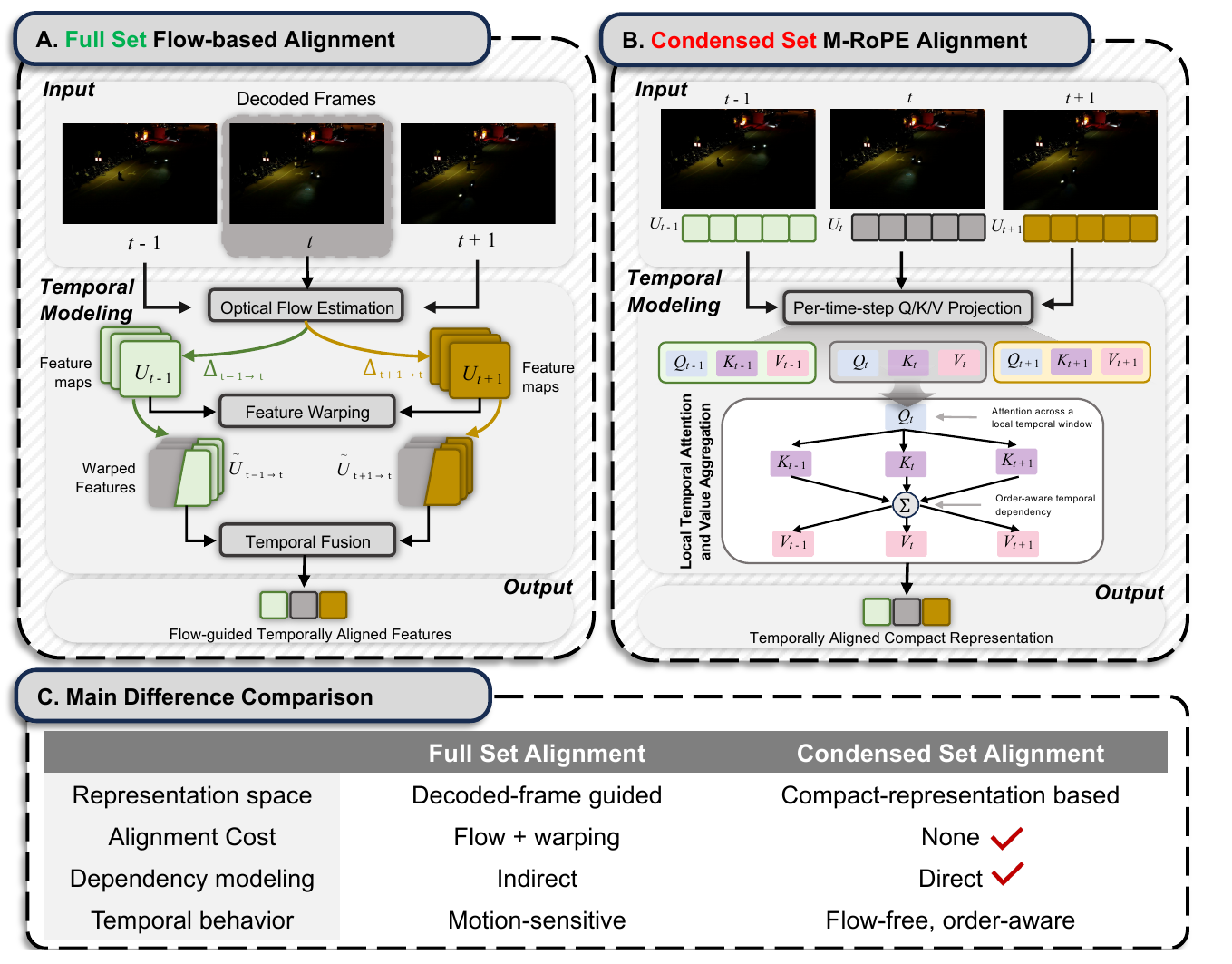}
    \vspace{-2.5em}
\caption{Comparison of temporal alignment paradigms under video fusion dataset condensation. 
Conventional flow-based alignment is not naturally suited to condensed synthetic clips 
% decoded from compact spatiotemporal representations
, whereas our method models order-aware temporal dependency in compact representation space.
% via per-time-step Q/K/V projection and M-RoPE attention.
}
\vspace{-2em}
\label{fig:temporal_alignment}
\end{figure}

%% file: section/2_relatedwork.tex
\section{Related Work}
\label{sec:related}

\textit{\textbf{Video Fusion.}}
Video fusion extends image fusion from static scenes to dynamic temporal sequences. 
While existing image fusion works, including CNN-~\cite{tang2018pixel,xu2020u2fusion}, Transformer-~\cite{ma2022swinfusion}, GAN-~\cite{ma2019fusiongan}, and diffusion-based~\citep{yi2024diff,zhao2023ddfm} approaches, have shown strong ability in multi-source information integration, they are primarily designed for single-frame settings and do not explicitly account for temporal dependency ~\citep{chen2024videomambasuite}. 
Directly applying image fusion models frame by frame to videos often causes flickering, motion discontinuity, and inter-frame inconsistency. 
Recent video fusion works~\citep{xie2024rcvs,tang2025videofusion,zhao2025unified,zhao2026mambavf,gong2025temcoco} therefore place greater emphasis on temporal modeling. UniVF~\citep{zhao2025unified} introduces multi-frame learning with flow-guided warping, but suffers from considerable computational overhead. MambaVF~\citep{zhao2026mambavf} alleviates this issue by replacing explicit motion alignment with state-space-based temporal modeling. TemCoCo~\citep{gong2025temcoco} further improves fused quality and temporal consistency through visual-semantic collaboration. 
However, these methods still rely on full-data training pipelines and original video representations, leaving compact condensed representations largely unexplored.

\vspace{0.3em}\noindent\textit{\textbf{Dataset Condensation.}} 
Existing dataset condensation methods can be broadly divided into three categories based on their matching objectives . In performance matching, DD~\citep{wang2018dataset} represents the meta-based line by learning synthetic samples through bilevel optimization, while RFAD~\citep{loo2022efficient} improves kernel-based performance matching with random feature approximation to reduce computational cost. In parameter matching, IDC~\citep{kim2022dataset} belongs to the single-step line and improves condensation efficiency through synthetic-data parameterization under limited storage, whereas MTT~\citep{cazenavette2022dataset} extends this idea to multi-step parameter matching by aligning training trajectories over multiple optimization steps, thereby alleviating the limitation of single-step matching, which only constrains local updates. In distribution matching, IDM~\citep{zhao2023improved} represents the single-layer line and improves naive feature distribution matching by addressing feature-number imbalance and unreliable embeddings for distance computation, while CAFE~\citep{wang2022cafe} further advances to multi-layer matching by aligning real and synthetic features across various scales, thus providing stronger global supervision and better discriminative preservation. Although these methods have achieved strong performance, most existing dataset condensation studies are still primarily developed and evaluated on image classification benchmarks~\citep{geng2023survey}, leaving limited exploration of multimodal video fusion, where condensed data should preserve not only cross-modal complementarity but also temporal dependency in compact spatiotemporal representations.

%% file: section/3_preliminary.tex
\section{Preliminaries}
\label{sec:preliminary}

\subsection{Problem Formulation}
\label{subsec:task_setup}

Given a pair of source videos from two modalities, denoted by $(\mathcal{V}_1,\mathcal{V}_2)$, each modality-specific video for $k\in\{1,2\}$ is written as
\begin{equation}
\mathcal{V}_k=\{\mathbf{I}_{k,t}\}_{t=1}^{T}.
\label{eq:video_sequence}
% \tag{1}
\end{equation}
Here, $\mathbf{I}_{k,t}\in\mathbb{R}^{C_k\times H\times W}$ denotes the $t$-th input frame, where $C_k$ is the channel number of modality $k$, $H\times W$ is the spatial resolution, and $T$ is the full video length. 

The goal of multi-source video fusion is to generate a fused video $\widehat{\mathcal{V}}_f=\{\widehat{\mathbf{I}}_{f,t}\}_{t=1}^{T}$ from $(\mathcal{V}_1,\mathcal{V}_2)$. Unlike static image fusion, multi-source video fusion must preserve both \textit{cross-modal complementarity} within each frame and \textit{temporal coherence} across neighboring frames. Accordingly, the task naturally involves two coupled aspects: intra-frame information aggregation and inter-frame temporal dependency modeling. The original training set is denoted by $\mathcal{T}$, which consists of paired source videos.

In data-efficient scenario, the fusion model is not applied to the full sequence at once. Instead, both training and inference are conducted on a local temporal window of length $L$. Therefore, $T$ denotes the full video length of a sample, whereas $L$ denotes the effective temporal context visible to the model at each step. This distinction is important because the subsequent temporal modeling is window-based rather than full-sequence based.

\subsection{Condensed Training Set}
\label{subsec:condensed_set}

Instead of directly training on the full dataset $\mathcal{T}$, we consider a condensed training set $\mathcal{S}$ containing $N_s$ learnable multi-source samples. Each condensed sample is parameterized as a pair of learnable compact spatiotemporal canvases $(\mathbf{G}_1,\mathbf{G}_2)$, where each modality-specific canvas for $k\in\{1,2\}$ satisfies
\begin{equation}
\mathbf{G}_k \in \mathbb{R}^{C_k \times \rho P_h \times \rho P_w}.
% \tag{2}
\label{eq:2}
\end{equation}
Here, $\rho$ denotes the grid size and $P_h\times P_w$ denotes the spatial size of each grid cell. Under this parameterization, one condensed sample contains $T_{\mathrm{cond}}=\rho^2$ compact frame units.
To recover video-like inputs from the compact canvases, we introduce a canvas decoder $\mathcal{U}$, which maps each $\mathbf{G}_k$ to a decoded clip:
\begin{equation}
\widetilde{\mathcal{V}}_k
=
\mathcal{U}(\mathbf{G}_k)
=
\{\widetilde{\mathbf{I}}_{k,t}\}_{t=1}^{T_{\mathrm{cond}}}.
% \tag{3}
\label{eq:3}
\end{equation}
Here, each decoded frame $\widetilde{\mathbf{I}}_{k,t}$ lies in $\mathbb{R}^{C_k\times H\times W}$, matching the channel dimensionality and spatial resolution of the original input frame. These decoded clips serve as the synthetic training source for subsequent fusion learning. Accordingly, $(\mathcal{S},\mathcal{U})$ defines the condensed-data branch used only during training, while only the final fusion model is retained at test time.

\subsection{Challenges under Condensed Representations}
\label{subsec:fusion_requirements}

Dataset condensation changes not only the amount of training data, but also the representation form on which fusion is learned. Instead of operating on natural video streams directly, the model is trained on decoded clips reconstructed from compact spatio-temporal canvases. 

Under this setting, video fusion faces the following three challenges.
\phantomsection
\label{challenge:temporal_mismatch}
\textbf{\textit{Challenge 1:}} Temporal modeling mismatch.
Conventional video fusion often relies on optical flow estimation and feature warping to establish cross-frame correspondence. However, such designs are developed for raw video frames or dense feature maps. In the condensed setting considered here, training is performed on decoded clips reconstructed from compact representations. As a result, conventional flow-based temporal alignment is not naturally matched to the representation form of condensed data.
\textbf{\textit{Challenge 2:}} Intra-frame information dilution.
In condensed representations, fusion-relevant cues are no longer expected to be uniformly distributed over all elements. Instead, they are more likely to concentrate in a sparse informative subset. Consequently, uniform aggregation can dilute the most useful information for fusion, making it difficult for the model to emphasize the elements that are truly critical.
\textbf{\textit{Challenge 3:}} Supervision preservation under data compression.
Effective video fusion depends on two coupled types of supervision: cross-modal complementarity within each frame and temporal coherence across neighboring frames. Once the training set is compressed into a small number of synthetic samples, both signals may be weakened simultaneously. Therefore, the last challenge is to preserve the fusion-relevant spatio-temporal supervision after substantial data reduction.

These three challenges show that dataset condensation for multi-source video fusion is not merely a data reduction problem. It also requires a fusion pipeline that remains compatible with condensed spatio-temporal representations.

%% file: section/4_methodology.tex
\section{STAR-VF}
\label{sec:method}

\input{Figure/Pipeline}

\subsection{Overview}
\label{subsec:method_overview}

As shown in Figure~\ref{fig:pipeline}, STAR-VF contains a condensed-data branch and a final fusion model. The condensed set is
\begin{equation}
\mathcal{S}=\{(\mathbf{G}_{1,s},\mathbf{G}_{2,s})\}_{s=1}^{N_s},\qquad
\mathbf{G}_{k,s}\in\mathbb{R}^{C_k\times\rho P_h\times\rho P_w},
\label{eq:compact_canvas}
\end{equation}
where $k\in\{1,2\}$. A modality-conditioned decoder $\mathcal{U}$ maps the $\rho\times\rho$ cells of each canvas, in row-major order $u=(a-1)\rho+b$, to $\widetilde{\mathcal{V}}_{k,s}=\mathcal{U}(\mathbf{G}_{k,s};k)=\{\widetilde{\mathbf{I}}_{k,s,u}\}_{u=1}^{\rho^2}$. Let $N_{\mathrm{real}}$ be the number of paired temporal units in the original set. The training-source ratio is
\(r_{\mathrm{cond}}=\frac{N_s\rho^2}{N_{\mathrm{real}}}.\)
The canvases are optimized variables rather than selected clips, and $\mathcal{U}$ converts their cells into full-resolution training inputs. We use $\rho=4$ and choose $N_s$ such that $r_{\mathrm{cond}}=1/16$. This ratio measures temporal training units rather than spatial downsampling.

Training has two stages. Stage I optimizes $\mathcal{S}$ and $\mathcal{U}$ by matching statistics extracted by a frozen proxy. Stage II fixes the condensed branch and trains $\mathcal{F}_{\Theta}$ on paired synthetic clips. For each target time $t$, the final model encodes a local window, models inter-frame dependency, aggregates informative elements from both modalities, and reconstructs the fused center frame. At inference, only $\mathcal{F}_{\Theta}$ is retained to process real video pairs.

\subsection{Flow-Free Inter-Frame Temporal Modeling}
\label{subsec:flowfree_temporal}

Decoded synthetic frames do not provide reliable motion fields for feature warping. We therefore model temporal dependency directly in representation space, allowing neighboring frames to interact without optical-flow estimation. This module uses the center frame as the query and the complete local window as temporal memory.

For a neighborhood $\mathcal{N}(t)$ of length $L$, the modality-specific encoder produces
\begin{equation}
\mathbf{X}_{k,\tau}=\mathcal{E}_k(\mathbf{I}_{k,\tau})\in\mathbb{R}^{J\times d},
\qquad \tau\in\mathcal{N}(t), k\in\{1,2\},
\label{eq:encoded_frame_representation}
\end{equation}
where $J$ is the number of tokens. Let $h$ be the number of heads and $d_h=d/h$. For head $m$,
\begin{equation}
\mathbf{Q}_{k,t}^{(m)}=\mathbf{X}_{k,t}\mathbf{W}_{Q}^{(m)},\quad
\mathbf{K}_{k,\tau}^{(m)}=\mathbf{X}_{k,\tau}\mathbf{W}_{K}^{(m)},\quad
\mathbf{V}_{k,\tau}^{(m)}=\mathbf{X}_{k,\tau}\mathbf{W}_{V}^{(m)},
\label{eq:qkv_temporal}
\end{equation}
with $\mathbf{W}_{Q}^{(m)},\mathbf{W}_{K}^{(m)},\mathbf{W}_{V}^{(m)}\in\mathbb{R}^{d\times d_h}$. For token $j$, M-RoPE uses $\mathbf{p}_{\tau,j}=(\tau,x_j,y_j)$ and $\Psi(\mathbf{v},\mathbf{p}_{\tau,j})=\mathbf{R}_{t}(\tau)\mathbf{R}_{x}(x_j)\mathbf{R}_{y}(y_j)\mathbf{v}$ to encode temporal order and spatial position. We concatenate the encoded keys and values over the window:
\begin{equation}
\overline{\mathbf{K}}_{k,t}^{(m)}=\operatorname*{Concat}_{\tau\in\mathcal{N}(t)}\Psi(\mathbf{K}_{k,\tau}^{(m)},\mathbf{p}_{\tau}),\quad
\overline{\mathbf{V}}_{k,t}^{(m)}=\operatorname*{Concat}_{\tau\in\mathcal{N}(t)}\mathbf{V}_{k,\tau}^{(m)},
\label{eq:temporal_memory}
\end{equation}
both in $\mathbb{R}^{LJ\times d_h}$. Temporal attention is normalized over all $LJ$ key tokens:
\begin{equation}
\mathbf{A}_{k,t}^{(m)}=\operatorname{Softmax}_{LJ}\!\left(
\frac{\Psi(\mathbf{Q}_{k,t}^{(m)},\mathbf{p}_{t})\overline{\mathbf{K}}_{k,t}^{(m)\top}}{\sqrt{d_h}}
\right),\qquad
\mathbf{H}_{k,t}^{(m)}=\mathbf{A}_{k,t}^{(m)}\overline{\mathbf{V}}_{k,t}^{(m)}.
\label{eq:temporal_attention}
\end{equation}
Joint normalization lets tokens from different frames compete while keeping the response scale independent of $L$. M-RoPE retains their temporal order and spatial location.

The enhanced center-frame representation is
\begin{equation}
\mathbf{Z}_{k,t}=\operatorname{Proj}\!\left[\operatorname{Concat}
(\mathbf{H}_{k,t}^{(1)},\ldots,\mathbf{H}_{k,t}^{(h)})\right]
\in\mathbb{R}^{J\times d}.
\label{eq:temporal_enhanced_representation}
\end{equation}

\input{table/1_comparison}

\input{Algorithm/Alg_1}

\subsection{Query-Driven Intra-Frame Aggregation}
\label{subsec:query_aggregation}

Fusion-relevant cues are not distributed uniformly across the $J$ tokens, so direct pooling may suppress localized or modality-specific details. For each modality, $M$ learnable queries $\mathbf{Q}^{\mathrm{agg}}_k\in\mathbb{R}^{M\times d}$ summarize the enhanced representation into aggregation slots. With $\mathbf{W}^{k}_{Q},\mathbf{W}^{k}_{K},\mathbf{W}^{k}_{V}\in\mathbb{R}^{d\times d}$, we compute
\begin{equation}
\widehat{\mathbf{Q}}_k=\mathbf{Q}^{\mathrm{agg}}_k\mathbf{W}^{k}_{Q},\qquad
\widehat{\mathbf{K}}_{k,t}=\mathbf{Z}_{k,t}\mathbf{W}^{k}_{K},\qquad
\widehat{\mathbf{V}}_{k,t}=\mathbf{Z}_{k,t}\mathbf{W}^{k}_{V},
\label{eq:qkv_aggregation}
\end{equation}
\begin{equation}
\mathbf{Y}_{k,t}=\operatorname{Softmax}_{J}\!\left(
\frac{\widehat{\mathbf{Q}}_k\widehat{\mathbf{K}}_{k,t}^{\top}}{\sqrt d}
\right)\widehat{\mathbf{V}}_{k,t}\in\mathbb{R}^{M\times d}.
\label{eq:query_driven_aggregation}
\end{equation}
The row-wise attention weights allow each slot to select a different subset of compact elements. This produces a structured summary for each modality while retaining non-uniform information selection.

The modality features are concatenated along the feature dimension and reconstructed by $\mathcal{D}:\mathbb{R}^{M\times2d}\rightarrow\mathbb{R}^{C_f\times H\times W}$:
\begin{equation}
\widehat{\mathbf{I}}_{f,t}=\mathcal{D}\!\left[\operatorname{Concat}_{d}
(\mathbf{Y}_{1,t},\mathbf{Y}_{2,t})\right].
\label{eq:reconstruction}
\end{equation}

\subsection{Condensed Set Learning and Final Fusion Training}
\label{subsec:two_stage_training}

\textit{\textbf{Stage I: Condensed Set Learning.}}
For each sampled real clip and decoded synthetic clip, a frozen proxy $\phi_{k,\ell}$~\cite{zhao2025unified} extracts multi-level features
\begin{equation}
\mathbf{F}^{r}_{k,\ell,n,t}=\phi_{k,\ell}(\mathbf{I}^{r}_{k,n,t}),\qquad
\mathbf{F}^{s}_{k,\ell,s,u}=\phi_{k,\ell}(\widetilde{\mathbf{I}}_{k,s,u}).
\label{eq:proxy_features}
\end{equation}
Following multi-level feature matching~\cite{wang2022cafe}, we use
\begin{equation}
\mathcal{L}_{\mathrm{cond}}
=
\underbrace{
\mathcal{D}^{16}_{\mathrm{mse}}
+
\lambda_{1}\mathcal{D}_{\mathrm{ch}}
+
\lambda_{2}
\left(
\mathcal{D}^{4}_{\mathrm{sp}}
+
\mathcal{D}^{8}_{\mathrm{sp}}
\right)
}_{\text{multi-level feature matching}}
+
\underbrace{
\lambda_{3}\mathcal{L}_{\mathrm{color}}
}_{\text{pixel-space matching}} .
\label{eq:condensation_loss}
\end{equation}
Here $\mathcal{D}^{16}_{\mathrm{mse}}$ matches mean $16\times16$ proxy feature maps, $\mathcal{D}_{\mathrm{ch}}$ matches channel mean and standard deviation, and $\mathcal{D}^{q}_{\mathrm{sp}}$ matches spatial mean and standard deviation on a $q\times q$ grid. Each term is summed over modalities and proxy levels after pooling over sampled clips and frames. $\mathcal{L}_{\mathrm{color}}$ applies the same moment matching in pixel space~\cite{gatys2016image,liu2025dcevo}, preventing feature matching from drifting toward implausible image statistics. Stage I therefore preserves representative spatial statistics in a compact paired source.

\vspace{0.3em}\noindent\textit{\textbf{Stage II: Final Fusion Model Training.}}
After fixing $(\mathcal{S},\mathcal{U})$, we train $\mathcal{F}_{\Theta}$ from scratch on local windows $\mathcal{W}_t$ sampled from the decoded clips:
\begin{equation}
\widehat{\mathbf{I}}_{f,t}=\mathcal{F}_{\Theta}(\mathcal{W}_t).
\label{eq:final_model_prediction}
\end{equation}
The fusion objective is
\begin{equation}
\mathcal{L}_{\mathrm{fuse}}
=
\mathcal{L}_{\mathrm{int}}
+
\mathcal{L}_{\mathrm{grad}}
+
\mathcal{L}_{\mathrm{ssim}}
+
\lambda_{4}\mathcal{L}_{\mathrm{temp}} .
\label{eq:fusion_loss}
\end{equation}
We use the intensity and gradient objectives of~\cite{xu2020u2fusion}, SSIM~\cite{wang2004image}, and the local temporal-consistency objective of~\cite{zhao2025unified} without modification. The first three terms supervise complementary source content in each fused frame, while $\mathcal{L}_{\mathrm{temp}}$ regularizes changes across adjacent predictions.

\input{Figure/qualitative}

%% file: Figure/Pipeline.tex
\begin{figure*}[t!]
    \centering
    \includegraphics[width=\textwidth]{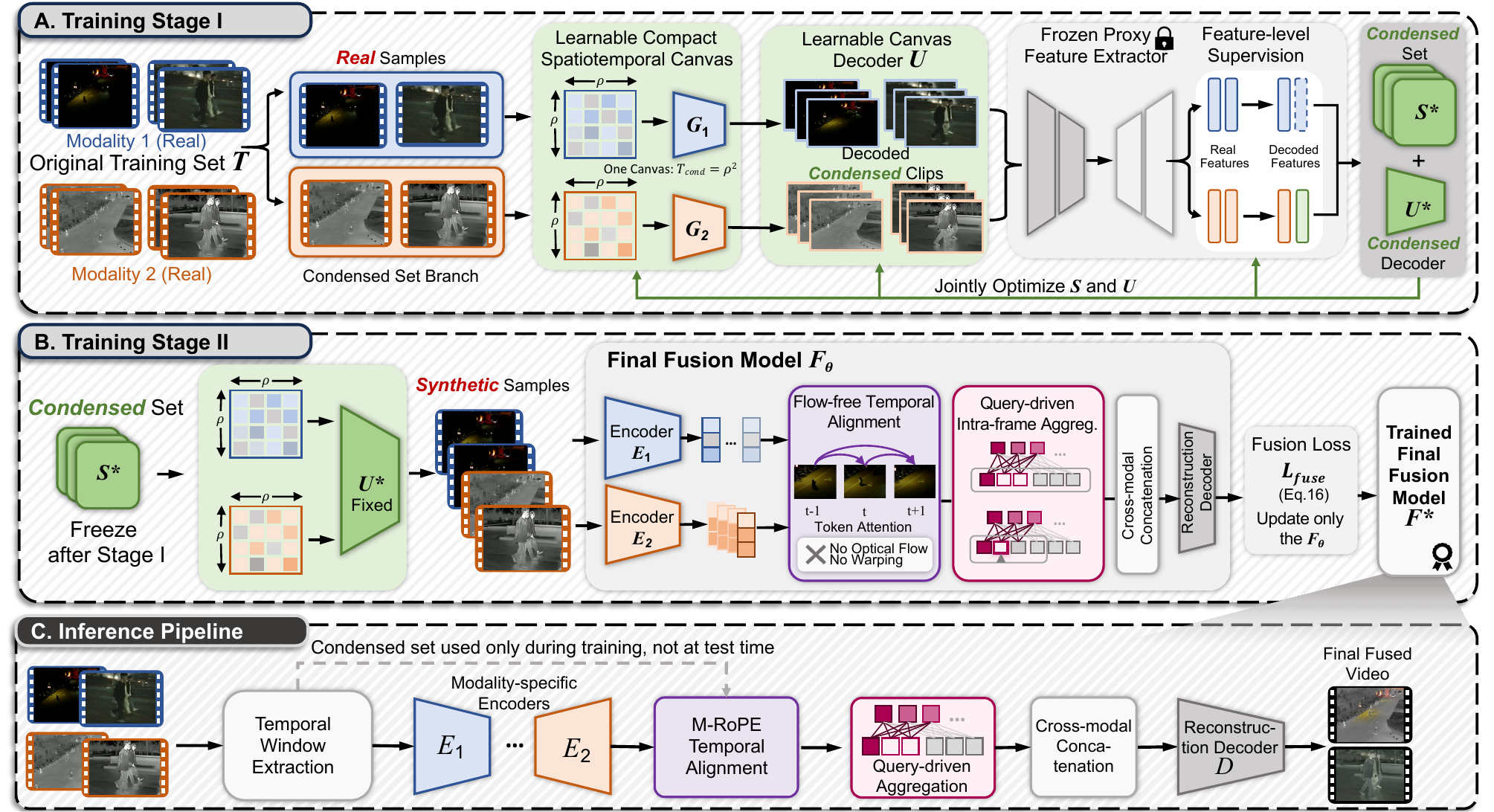}
    \vspace{-2.5em}
    \caption{Overview of the proposed STAR-VF framework.
    In Stage I, learnable compact spatio-temporal canvases and the canvas decoder $U$ are jointly optimized under feature-level supervision from a frozen proxy feature extractor, yielding the learned condensed set $S^*$ and decoder $U^*$.
    In Stage II, the learned condensed set is decoded into synthetic temporal samples to train the final fusion model $F_{\theta}$ with flow-free temporal alignment and query-driven intra-frame aggregation.
    During inference, only the trained fusion model is used, without the condensed set or proxy model.}
    \label{fig:pipeline}
\end{figure*}

%% file: table/1_comparison.tex
\begin{table*}[h]
\centering
\footnotesize
\caption{
Quantitative comparison on IVF (VTMOT), MVF (Harvard), MEF
(YouTube-HDR), and MFF (DAVIS).
The training-data ratio $r_{\mathrm{train}}$ denotes the fraction of
the original temporal training units used to train each method.
STAR-VF uses only approximately 1/16 of the original training source, yet
achieves fusion quality comparable to the full-data baselines.
The best and second-best results are highlighted in \colorbox{red!20}{\textbf{red}} and \colorbox{blue!15}{\textbf{blue}}, respectively.
}
\vspace{-1.5em}
\label{tab:all_video_fusion}

\setlength{\tabcolsep}{2pt}{
\begin{tabular}{l c ccccc ccccc ccccc ccccc}
\toprule
\multirow{2}{*}{\textbf{Method}}
& \multirow{2}{*}{\shortstack{\textbf{Train}\\\textbf{Ratio}}}
& \multicolumn{5}{c}{\textbf{IVF (VTMOT~\cite{zhu2025visible})}}
& \multicolumn{5}{c}{\textbf{MVF (Harvard~\cite{summers2003harvard})}}
& \multicolumn{5}{c}{\textbf{MEF (YouTube-HDR~\cite{wang2024youtube})}}
& \multicolumn{5}{c}{\textbf{MFF (DAVIS~\cite{pont20172017})}} \\

\cmidrule(lr){3-7}
\cmidrule(lr){8-12}
\cmidrule(lr){13-17}
\cmidrule(lr){18-22}

& & CC$\uparrow$ & EN$\uparrow$ & SSIM$\uparrow$
& BiSWE$\downarrow$ & MS2R$\downarrow$
& CC$\uparrow$ & EN$\uparrow$ & SSIM$\uparrow$
& BiSWE$\downarrow$ & MS2R$\downarrow$
& CC$\uparrow$ & EN$\uparrow$ & SSIM$\uparrow$
& BiSWE$\downarrow$ & MS2R$\downarrow$
& CC$\uparrow$ & EN$\uparrow$ & SSIM$\uparrow$
& BiSWE$\downarrow$ & MS2R$\downarrow$ \\
\midrule

UniVF~\citep{zhao2025unified}
& 1
& \cellcolor{red!20}0.65 & \cellcolor{red!20}6.81
& \cellcolor{red!20}0.65 & 3.95 & \cellcolor{red!20}0.35
& \cellcolor{blue!15}0.78 & 4.27 & 0.75 & 29.92
& \cellcolor{red!20}1.23
& \cellcolor{red!20}0.96 & 7.25 & \cellcolor{red!20}0.65
& \cellcolor{red!20}6.44 & \cellcolor{blue!15}0.34
& \cellcolor{red!20}0.98 & 7.41 & \cellcolor{blue!15}0.90
& \cellcolor{blue!15}6.04 & \cellcolor{red!20}1.11 \\

VideoFusion~\citep{tang2025videofusion}
& 1
& 0.62 & 6.59 & 0.62 & \cellcolor{red!20}3.14 & 0.59
& \cellcolor{blue!15}0.78 & \cellcolor{blue!15}4.91
& 0.69 & \cellcolor{red!20}19.71 & 1.31
& \cellcolor{blue!15}0.91 & 6.82 & \cellcolor{blue!15}0.58
& \cellcolor{blue!15}9.98 & 1.82
& 0.96 & 7.34 & 0.82 & 9.07 & 6.13 \\

TemCoCo~\citep{gong2025temcoco}
& 1
& 0.63 & 6.68 & 0.60 & 6.26 & 0.59
& 0.77 & \cellcolor{red!20}5.40 & 0.24 & 28.41 & 1.57
& 0.90 & \cellcolor{red!20}7.34 & 0.53 & 11.62 & 1.70
& \cellcolor{blue!15}0.97 & \cellcolor{red!20}7.45
& 0.84 & 12.34 & 4.08 \\

MambaVF~\cite{zhao2026mambavf}
& 1
& - & - & 0.65 & 3.91 & 0.36
& - & - & \cellcolor{red!20}0.76 & 29.61 & 1.27
& - & - & - & 6.96 & \cellcolor{red!20}0.16
& - & - & - & 8.29 & - \\

STAR-VF (ours)
& $\sim$1/16
& \cellcolor{blue!15}0.64 & \cellcolor{blue!15}6.73
& \cellcolor{red!20}0.65 & \cellcolor{blue!15}3.91
& \cellcolor{red!20}0.35
& \cellcolor{red!20}0.80 & 4.37 & \cellcolor{red!20}0.76
& \cellcolor{blue!15}28.36 & \cellcolor{blue!15}1.26
& \cellcolor{blue!15}0.91 & \cellcolor{blue!15}7.28
& 0.55 & 10.23 & 1.28
& 0.96 & \cellcolor{blue!15}7.42 & \cellcolor{red!20}0.91
& \cellcolor{red!20}5.53 & \cellcolor{blue!15}1.18 \\

\bottomrule
\end{tabular}
}
\end{table*}

%% file: Algorithm/Alg_1.tex
\begin{algorithm}[t]
\caption{Two-Stage Training Pipeline for Video Fusion}
\label{alg:overall_framework}
\small
\begin{algorithmic}[1]
\Require training set $\mathcal{T}$, condensed set size $N_s$, and window length $L$
\Ensure $(\mathcal{S}^\ast,\mathcal{U}^\ast,\mathcal{F}_{\Theta}^\ast)$
\Statex

\State Initialize $\mathcal{S}$ and $\mathcal{U}$
\Statex {\bfseries Stage I: Condensation}
\For{each iteration}
    \State Sample a real clip pair from $\mathcal{T}$
    \Comment{real supervision source}
    \State Sample $s\in\{1,\ldots,N_s\}$ and decode
    $\widetilde{\mathcal{V}}_{k,s}
    \gets \mathcal{U}(\mathbf{G}_{k,s};k)$ for $k\in\{1,2\}$
    \Comment{synthetic clips from compact canvases}
    \State Compute $\mathcal{L}_{\mathrm{cond}}$ and update
    $\mathcal{S},\mathcal{U}$
    \Comment{learn condensed branch}
\EndFor
\State $(\mathcal{S}^\ast,\mathcal{U}^\ast)
\gets(\mathcal{S},\mathcal{U})$
\Comment{freeze for Stage II}

\Statex
\State Initialize $\mathcal{F}_{\Theta}$
\Statex {\bfseries Stage II: Fusion Training}
\For{each iteration}
    \State Decode synthetic clips from
    $(\mathcal{S}^\ast,\mathcal{U}^\ast)$
    \State Construct $\mathcal{W}_t$ with length $L$
    \Comment{local temporal window}
    \State Predict
    $\widehat{\mathbf{I}}_{f,t}
    \gets \mathcal{F}_{\Theta}(\mathcal{W}_t)$
    \Comment{center-frame fusion}
    \State Compute $\mathcal{L}_{\mathrm{fuse}}$ and update $\Theta$
    \Comment{train final fusion model}
\EndFor
\State $\mathcal{F}_{\Theta}^\ast
\gets \mathcal{F}_{\Theta}$

\State \Return
$(\mathcal{S}^\ast,\mathcal{U}^\ast,\mathcal{F}_{\Theta}^\ast)$
\end{algorithmic}
\end{algorithm}

%% file: Figure/qualitative.tex
\begin{figure*}[h]
    \centering
    \includegraphics[width=\textwidth]{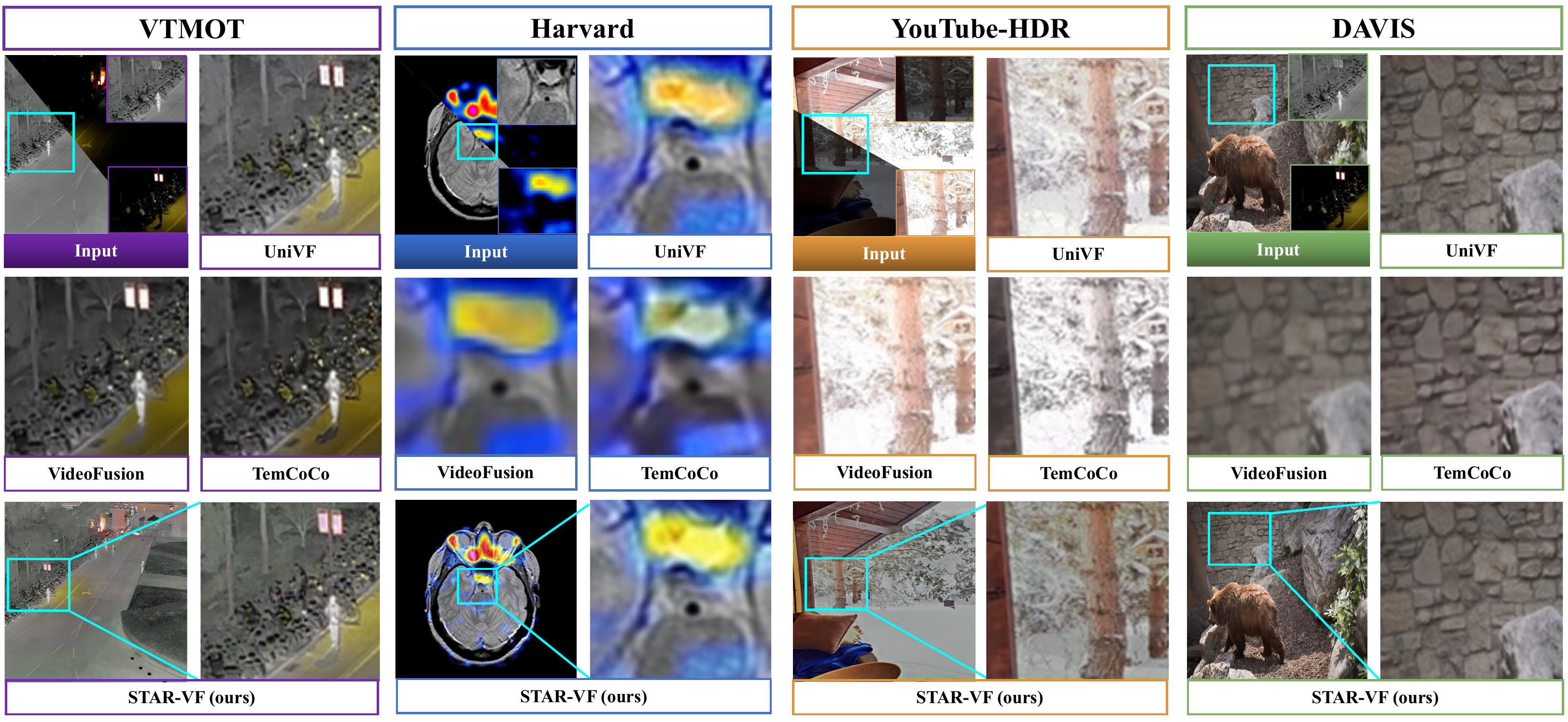}
    \vspace{-2.5em}
    \caption{
    Qualitative comparison on four representative benchmarks. For each example, we show the input pair, our method, the full-size variant trained on the original data, and competing video fusion baselines (\textit{VideoFusion}, \textit{UniVF}, and \textit{TemCoCo}).
    The cyan boxes indicate enlarged local regions for detailed comparison. 
    % Compared with the baselines, our method preserves clearer structures and more faithful salient content, while remaining visually close to the full-size model.
    }
    % \vspace{-0.5em}
    \label{fig:qualitative_comparison}
\end{figure*}

%% file: section/5_experiments.tex
\section{Experiments}
\label{sec:exp}

\input{Figure/qualitative_video}

\subsection{Experimental Setup}

\textit{\textbf{Datasets and Metrics.}}
We evaluate STAR-VF on four video-fusion tasks: infrared-visible fusion (IVF) on VTMOT~\citep{zhu2025visible}, medical video fusion (MVF) on Harvard~\citep{summers2003harvard}, multi-exposure fusion (MEF) on YouTube-HDR~\citep{wang2024youtube}, and multi-focus fusion (MFF) on DAVIS~\citep{pont20172017}. These datasets are used under their corresponding fusion tasks following the UniVF protocol, without additional custom pairing, alignment, or split construction. We compare with the dedicated video-fusion methods UniVF~\citep{zhao2025unified}, VideoFusion~\citep{tang2025videofusion}, TemCoCo~\citep{gong2025temcoco}, and MambaVF~\citep{zhao2026mambavf}.

We report the image-level metrics CC, EN, and SSIM used in IVIF-ZOO~\citep{liu2024infrared}, together with the video-level temporal metrics BiSWE and MS2R from VF-Bench~\citep{zhao2025unified}. Higher CC, EN, and SSIM and lower BiSWE and MS2R indicate better performance. Efficiency is measured by parameters, FLOPs, latency, and peak GPU memory.

\vspace{0.3em}\noindent\textit{\textbf{Implementation Details.}}
In Stage I, compact canvases are initialized by sampling and spatially interpolating real frames. A frozen UniVF model provides proxy features, and the condensed branch is optimized for 10{,}000 iterations using Adam~\citep{kingma2014adam} with $(\beta_1,\beta_2)=(0.9,0.999)$ and a learning rate of $10^{-2}$. In Stage II, the condensed branch is fixed and a fusion model with $C=32$ and four Transformer blocks is trained from scratch for 20{,}000 iterations. We use $M=8$ aggregation queries and the centered 3-frame window $\mathcal{N}(t)=\{t-1,t,t+1\}$. AdamW~\cite{loshchilov2017decoupled}, a cosine schedule from $2{\times}10^{-4}$ to $10^{-6}$, and a batch size of 8 are used. 
We empirically set $\lambda_1$, $\lambda_2$, $\lambda_3$ and $\lambda_4$ to
0.5, 0.1, 0.1 and 0.5, respectively.

\subsection{Main Results}

\noindent\textit{\textbf{Quantitative Results.}}
Table~\ref{tab:all_video_fusion} shows that STAR-VF, trained with the dataset-specific synthetic budgets in Table~\ref{tab:condensation_budget}, remains competitive with full-data methods. It ranks first in CC and SSIM on MVF and in SSIM and BiSWE on MFF, while remaining close to the best MS2R, showing that a compact training source retains useful per-frame fusion cues and temporal regularity.

\vspace{0.3em}\noindent\textit{\textbf{Qualitative Results.}}
Figure~\ref{fig:qualitative_comparison} shows that STAR-VF preserves salient structures and complementary details across all four tasks, including pedestrian contours on VTMOT, tissue boundaries on Harvard, highlight and shadow details on YouTube-HDR, and focus transitions on DAVIS. The enlarged regions expose local structural and brightness differences. In Figure~\ref{fig:framewise_comparison}, STAR-VF also exhibits fewer brightness fluctuations and more consistent colors than the compared methods.

\begin{table}[t]
  \centering
  \footnotesize
  \caption{Training-source budgets used by STAR-VF.}
  \vspace{-1.5em}
  \label{tab:condensation_budget}
  \begin{tabular*}{\linewidth}{@{\extracolsep{\fill}}lccrrrr@{}}
    \toprule
    \textbf{Dataset} & \textbf{Task} & \textbf{Train seqs} & \textbf{$N_{\mathrm{real}}$} & \textbf{$N_s$} & \textbf{$N_{\mathrm{cond}}$} & \textbf{$r_{\mathrm{cond}}$} \\
    \midrule
    VTMOT       & IVF & 75  & 22{,}500 & 88  & 1{,}408 & $1/16.0$ \\
    Harvard     & MVF & 49  & 1{,}308  & 6   & 96      & $1/13.6$ \\
    YouTube-HDR & MEF & 450 & 66{,}595 & 261 & 4{,}176 & $1/15.9$ \\
    DAVIS       & MFF & 120 & 8{,}195  & 32  & 512     & $1/16.0$ \\
    \bottomrule
  \end{tabular*}
\end{table}

\input{table/2_effiency}

\vspace{0.3em}\noindent\textit{\textbf{Training-Source Efficiency.}}
Table~\ref{tab:condensation_budget} reports the actual budget on each benchmark, where $N_{\mathrm{cond}}=N_sT_{\mathrm{cond}}$. All settings use $\rho=4$, $P_h\times P_w=64\times64$, and $T_{\mathrm{cond}}=16$. STAR-VF reduces the training source to approximately $1/16$ of the original temporal units on three datasets and $1/13.6$ on Harvard, substantially lowering the data required for fusion training.

\vspace{0.3em}\noindent\textit{\textbf{Computational Efficiency.}}
Table~\ref{tab:flops_new} compares the deployed fusion architectures; the condensed branch and proxy are absent at inference. Relative to UniVF, STAR-VF reduces parameters from $9.16$M to $0.32$M. At $256{\times}256$, FLOPs and latency decrease from $934.7$G and $531.4$ ms to $184.0$G and $110.5$ ms; at $480{\times}854$, they decrease from $6139$G and $820.7$ ms to $1066.6$G and $468.5$ ms. These inference gains arise from the final lightweight architecture and are distinct from the reduction in training-data budget.

\vspace{0.3em}\noindent\textit{\textbf{Frame-Wise Consistency.}}
We extract luminance trajectories from aligned regions of interest and apply min--max normalization to remove scale differences. Figure~\ref{fig:framewise_consistency} shows that STAR-VF follows the shared temporal trend with smaller local deviations, indicating a more stable fused appearance.

\input{Figure/condensed_dataset}
\input{Figure/alignmentanalysis}

\subsection{Ablation Studies}

\textit{\textbf{Effect of Flow-Free Temporal Modeling.}}
We replace VideoFusion's RAFT-based alignment and warping with M-RoPE attention while keeping the full training set and fusion setting unchanged. Table~\ref{tab:ablation1} shows comparable fusion quality and temporal consistency on IVF and MFF, indicating that explicit flow estimation can be removed without performance loss while preserving effective temporal interactions across adjacent frames.

\vspace{0.3em}\noindent\textit{\textbf{Effect of Condensation Grid Size.}}
Table~\ref{tab:ablation2} shows that the $4{\times}4$ grid achieves the lowest MS2R ($0.35$) and lower BiSWE ($3.91$), providing the best balance between compactness and temporal stability.

\vspace{0.3em}\noindent\textit{\textbf{Attribution to Learned Condensation.}}
Under the same architecture and $1/16$ budget on VTMOT, random compact sources achieve $5.64{\pm}0.79$ EN, $0.48{\pm}0.12$ CC, and $0.51{\pm}0.08$ SSIM, compared with $6.73$, $0.64$, and $0.65$ for learned condensation. This consistent improvement under an identical data budget confirms that the gains come from optimized condensation rather than arbitrary data reduction or source-size effects.

\vspace{0.3em}\noindent\textit{\textbf{Proxy Robustness and Transferability.}}
Table~\ref{tab:proxy_robustness} shows that condensed sources generated with ResNet-18 and VGG-16 remain effective for both STAR-VF and UniVF, demonstrating robustness to proxy choice and transferability across fusion models.

\input{table/3_ablation}
\input{table/proxy}

%% file: Figure/qualitative_video.tex
\begin{figure}[t]
    \centering
    \includegraphics[width=\linewidth]{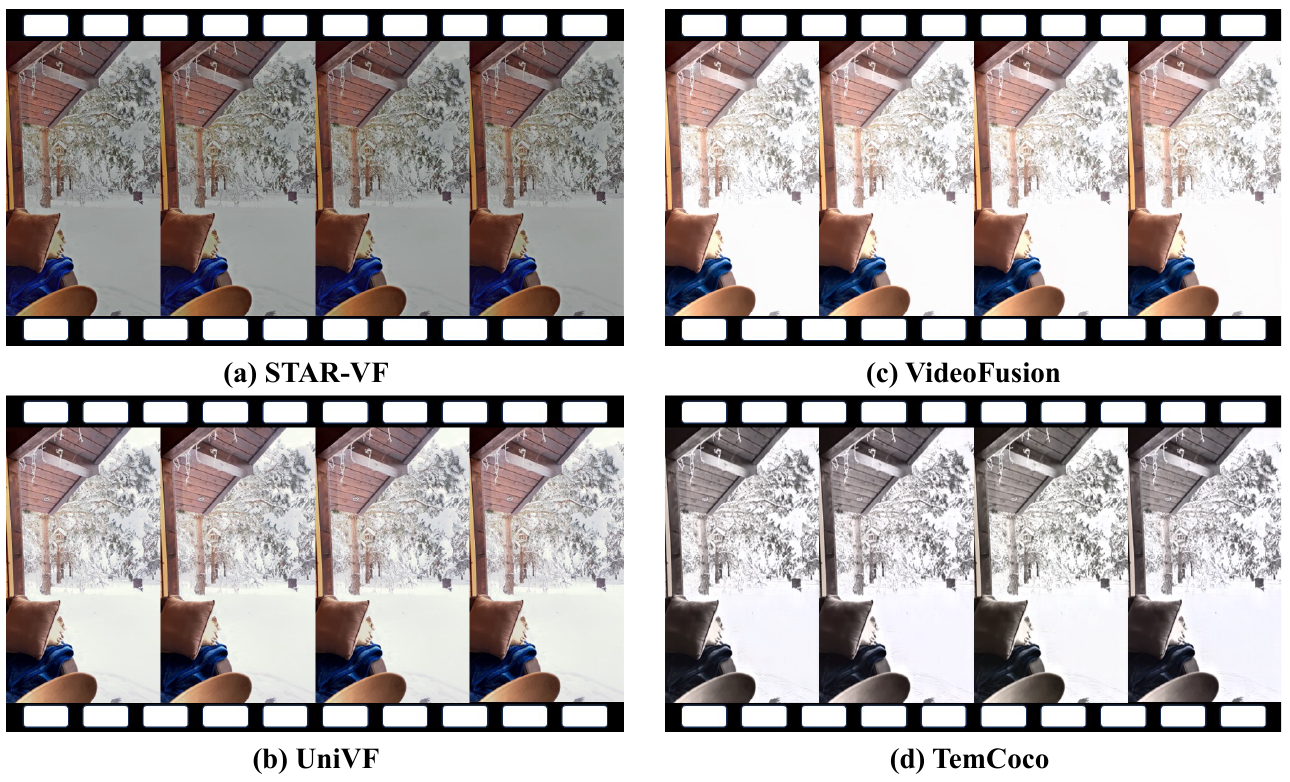}
    \vspace{-2.5em}
    \caption{Frame-by-frame qualitative comparison on a representative multi-source video fusion sequence. }
    \label{fig:framewise_comparison}
    % \vspace{-0.5em}
\end{figure}

%% file: table/2_effiency.tex
\begin{table}[t]
  \centering
  \footnotesize
  \caption{Computational efficiency comparison.}
  \vspace{-1.5em}
  \label{tab:flops_new}
  % \resizebox{\linewidth}{!}{
  \setlength{\tabcolsep}{4.5pt}{
  \begin{tabular}{l ccccc}
    \toprule
    \textbf{Model} & \textbf{Resolution} & \textbf{Params} & \textbf{GFLOPs} & \textbf{Latency} & \textbf{Memory} \\
    \midrule
    UniVF~\citep{zhao2025unified} & 256$\times$256 & 9.16M & 934.7 & 531.4ms & 696.8MB \\
    Condensed (ours) & 256$\times$256 & 0.32M & 184.0 & 110.5ms & 696.8MB \\
    \midrule
    UniVF~\citep{zhao2025unified} & 480$\times$854 & 9.16M & 6139.1 & 820.7ms & 4110.5MB \\
    Condensed (ours) & 480$\times$854 & 0.32M & 1066.6 & 468.5ms & 4110.5MB \\
    \bottomrule
  \end{tabular}}
\end{table}

%% file: Figure/condensed_dataset.tex
\begin{figure}[t]
    \centering
    \includegraphics[width=\linewidth]{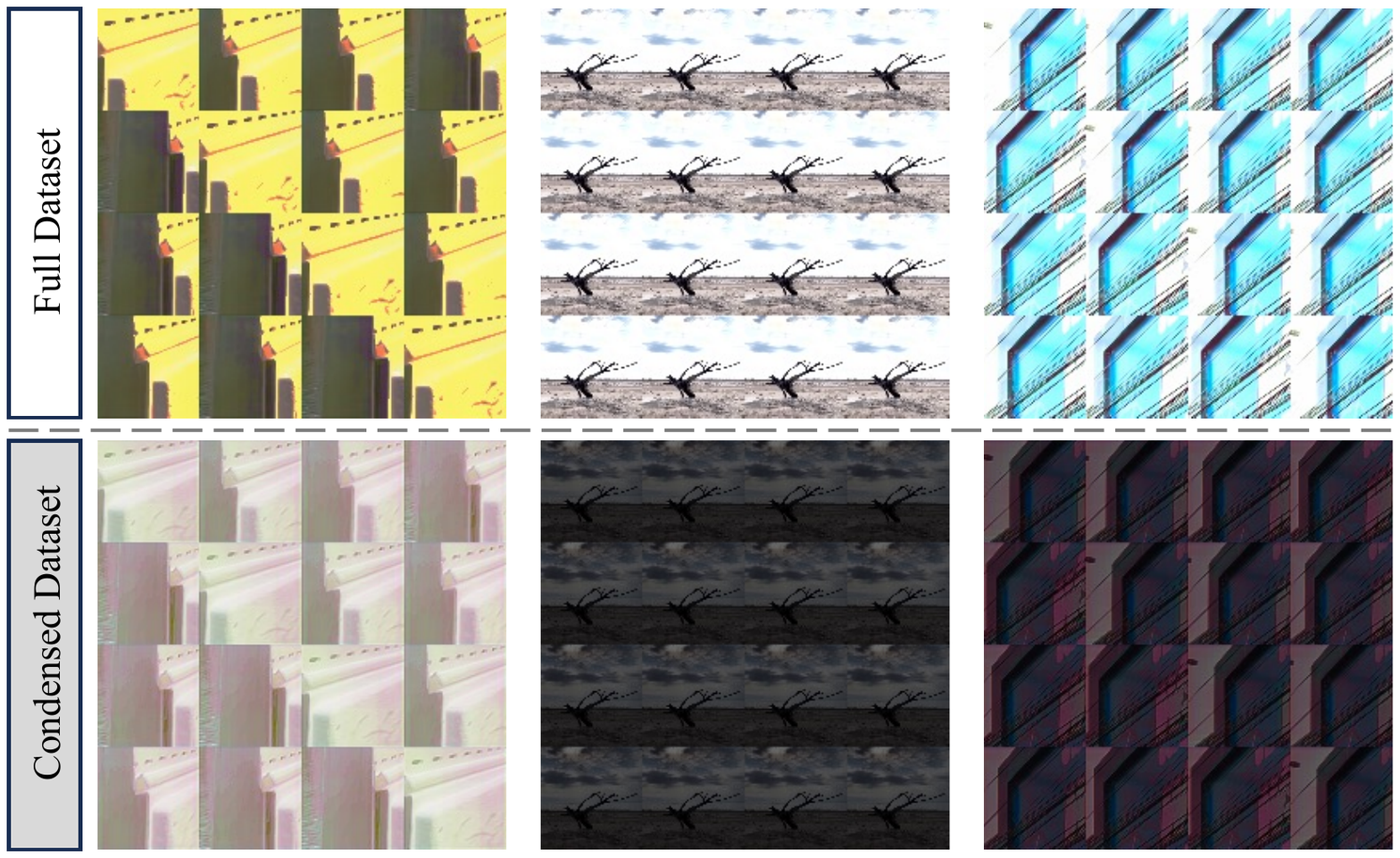}
    \vspace{-4em}
    \caption{
    Visual comparison between the full dataset and the learned condensed dataset.
    }
        \vspace{-1em}
    \label{fig:condensed_dataset_visualization}
\end{figure}

%% file: Figure/alignmentanalysis.tex
\begin{figure}[t]
    \centering
    \includegraphics[width=\linewidth]{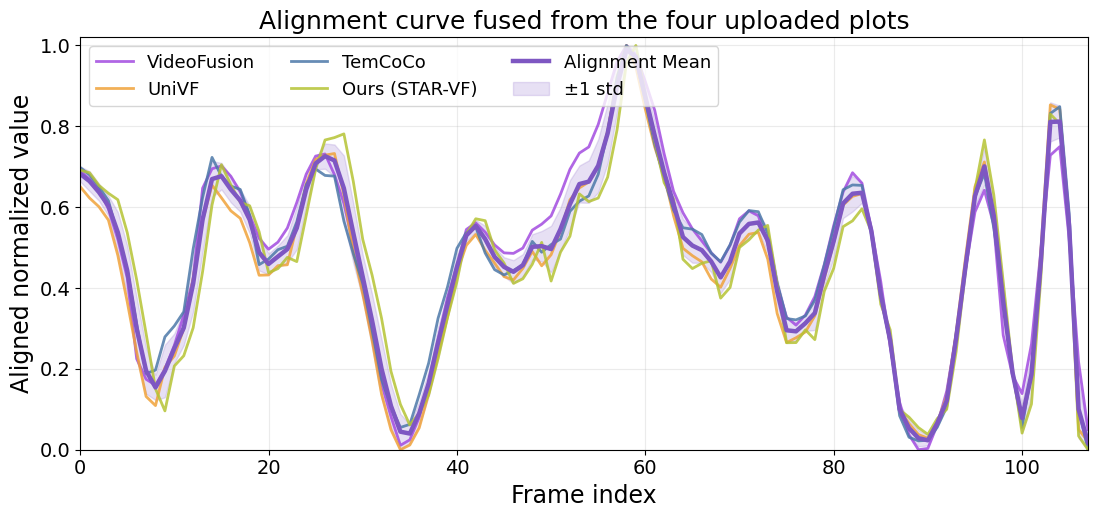}
    \vspace{-2.5em}
    \caption{Alignment consistency comparison across video fusion methods, evaluated by frame-wise aligned ROI luminance trajectories after min-max normalization }
    \label{fig:framewise_consistency}
        \vspace{-1em}
\end{figure}

%% file: table/3_ablation.tex
\begin{table}[t]
\centering
\footnotesize
\caption{Ablation of flow-free temporal modeling on IVF.}
  \vspace{-1.5em}
\label{tab:ablation1}
\setlength{\tabcolsep}{8pt}{
\begin{tabular}{lccccc}
\toprule
\textbf{Method} &  \textbf{CC$\uparrow$} &  \textbf{EN$\uparrow$} &  \textbf{SSIM$\uparrow$} &  \textbf{BiSWE$\downarrow$} &  \textbf{MS2R$\downarrow$} \\
\midrule
UniVF~\citep{zhao2025unified}
& 0.65 & 6.81 & 0.65 & 3.95 & 0.35 \\
w/o condensation
& 0.64 & 6.73 & 0.64 & 3.91 & 0.36 \\
\midrule
Ours (w/o RAFT)
& 0.64 & 6.73 & 0.65 & 3.91 & 0.35 \\
\bottomrule
\end{tabular}}
\end{table}

\begin{table}[t]
\centering
\footnotesize
\caption{Ablation on condensation grid size on IVF.}
\vspace{-1.5em}
\label{tab:ablation2}
\setlength{\tabcolsep}{6.25pt}{
\begin{tabular}{lc ccccc}
\toprule
\textbf{Grid} & \textbf{\#Frames} &  \textbf{CC$\uparrow$} &  \textbf{EN$\uparrow$} &  \textbf{SSIM$\uparrow$} &  \textbf{BiSWE$\downarrow$} &  \textbf{MS2R$\downarrow$} \\
\midrule
1$\times$1 (full) &  1 & 0.64 & 6.73 & 0.64 & 3.91 & 0.36 \\
2$\times$2 &  4 & 0.90 & 7.25 & 0.87 & 5.07 & 0.36 \\
3$\times$3 &  9 & 0.89 & 7.82 & 0.88 & 5.10 & 0.36 \\
4$\times$4 (ours) & 16
  & \textbf{0.64} & \textbf{6.73} & \textbf{0.65}
  & \textbf{3.91} & \textbf{0.35} \\
5$\times$5 & 25 & 0.90 & 6.62 & 0.88 & 5.17 & 0.36 \\
\bottomrule
\end{tabular}}
\end{table}

%% file: table/proxy.tex
\begin{table}[t]
    \centering
    \caption{Proxy robustness and downstream transferability on VTMOT. ``Full'' denotes UniVF trained on the complete real set; all other rows use a $1/16$ condensed source.}
  \vspace{-1.5em}
    \label{tab:proxy_robustness}
    \footnotesize
    \setlength{\tabcolsep}{4.45pt}
    \begin{tabular}{llccccc}
        \toprule
        \textbf{Downstream} & \textbf{Source/Proxy}
        & \textbf{EN$\uparrow$} & \textbf{CC$\uparrow$} & \textbf{SSIM$\uparrow$}
        & \textbf{BiSWE$\downarrow$} & \textbf{MS2R$\downarrow$} \\
        \midrule
        UniVF & Full
        & 6.814 & 0.654 & 0.652 & 3.950 & 0.353 \\
        UniVF & Default
        & 6.370 & 0.625 & 0.626 & 5.505 & 0.379 \\
        UniVF & ResNet-18
        & 6.320 & 0.601 & 0.626 & 5.101 & 0.382 \\
        UniVF & VGG-16
        & 6.510 & 0.660 & 0.628 & 5.205 & 0.375 \\
        \midrule
        STAR-VF & Default
        & 6.731 & 0.641 & 0.648 & 3.920 & 0.356 \\
        STAR-VF & ResNet-18
        & 6.720 & 0.591 & 0.635 & 3.586 & 0.355 \\
        STAR-VF & VGG-16
        & 6.709 & 0.643 & 0.646 & 3.445 & 0.354 \\
        \bottomrule
    \end{tabular}
\end{table}

%% file: section/6_conclusion.tex
\section{Conclusion}
\label{sec:conclusion}
In this paper, we introduced STAR-VF, the first spatio-temporal representation condensation framework for data-efficient video fusion. Across four representative tasks, STAR-VF achieves performance comparable to full-data-trained methods using only a $1/16$ training-source budget, demonstrating that compact synthetic representations can preserve fusion-relevant spatial and temporal cues. Controlled comparisons further show that the improvement arises from learned condensation rather than random compact construction or model-size reduction. Experiments with different proxy networks also indicate that the condensed supervision is transferable and not tied to a specific feature extractor or downstream model. Since the condensed branch and proxy are used only during training, inference operates directly on real multi-source videos using the final fusion model. Nevertheless, performance remains bounded by the capacity of the condensed source, and a fixed budget may become insufficient as the diversity of original data increases.

%% file: section/x_appendix.tex
% Appendix-local numbering and hyperlink helpers.
\makeatletter
\@addtoreset{equation}{section}
\makeatother
\renewcommand{\theequation}{\thesection-\arabic{equation}}
\renewcommand{\theHequation}{\thesection.\arabic{equation}}

\newcommand{\appsecref}[1]{Section~\ref{#1}}
\newcommand{\appsubsecref}[1]{Section~\ref{#1}}
\newcommand{\appeqref}[1]{Eq.~\eqref{#1}}
\newcommand{\appfigref}[1]{Figure~\ref{#1}}
\newcommand{\apptabref}[1]{Table~\ref{#1}}

\section*{\huge Appendix}

\startcontents[app]
\section*{Table of Contents}
\printcontents[app]{l}{1}[2]{}

\newpage
\section{Problem Setup and Compact Representation}
\label{app:setup}

This section expands the formulation in \mainsecref{sec:preliminary} and the condensed-set definition in \mainsecref{subsec:method_overview}. We distinguish the temporal extent of a video from the context used for one prediction, explain why video-fusion condensation differs from static condensation, and make the compact training source and its decoder explicit.

\subsection{Full Sequences and Local Fusion Windows}
\label{app:local_formulation}

For modality $k\in\{1,2\}$, a paired input sequence and the corresponding fused output are
\begin{equation}
\mathcal{V}_k=\{\mathbf{I}_{k,t}\}_{t=1}^{T},
\qquad
\widehat{\mathcal{V}}_f=\{\widehat{\mathbf{I}}_{f,t}\}_{t=1}^{T},
\label{eq:app_sequence}
\end{equation}
where $\mathbf{I}_{k,t}\in\mathbb{R}^{C_k\times H\times W}$ and $T$ is the sequence length. The target is defined over all $T$ frames, but the model predicts each center frame from a neighborhood $\mathcal{N}(t)$ of length $L$:
\begin{equation}
\mathcal{W}_t=\{(\mathbf{I}_{1,\tau},\mathbf{I}_{2,\tau})\}_{\tau\in\mathcal{N}(t)},
\quad |\mathcal{N}(t)|=L,
\quad
\widehat{\mathbf{I}}_{f,t}=\mathcal{F}_{\Theta}(\mathcal{W}_t).
\label{eq:app_local_operator}
\end{equation}
Sliding the operator in \appeqref{eq:app_local_operator} over $t=1,\ldots,T$ produces the full output in \appeqref{eq:app_sequence}. Thus, $T$ is a property of the data sequence, whereas $L$ is the temporal receptive field of one fusion decision. This distinction also separates output scope from computational scope: STAR-VF generates a full video while using a bounded amount of context at each time step.

The local formulation matches both training sources. For real videos, $\mathcal{W}_t$ is sampled from the original paired sequences. For condensed data, the same construction is applied to ordered frames decoded from a compact canvas. The final model therefore sees the same input type in Stage~II and at inference; only the source of the frames changes. Boundary positions can be handled by the same padding or replication rule used by the implementation, without changing the operator itself.

Local windows also make the temporal supervision scale explicit. The condensed branch does not need to reproduce an unrestricted video history before the fusion model can learn temporal regularity. It needs to provide enough ordered support to form the neighborhoods consumed by $\mathcal{F}_{\Theta}$. The relation between $L$ and the useful temporal range is analyzed in \appsubsecref{app:local_scope}.

\subsection{Why Video-Fusion Condensation Requires a Structured Source}
\label{app:video_condensation_challenges}

Static condensation usually represents each synthetic example as an independent image and preserves supervision associated with one spatial input. Video fusion adds two coupled axes. Within a frame, the two modalities contribute different and spatially non-uniform evidence; across frames, the output should remain stable while preserving genuine scene changes. A condensed source must therefore retain cross-modal complementarity and temporal organization at the same time. Compressing each frame independently can preserve local appearance yet discard the ordering needed to construct informative temporal neighborhoods.

The representation used for temporal learning also changes after condensation. Real videos contain natural pixel-space continuity, whereas decoded synthetic clips are optimized to preserve downstream supervision under a limited budget. Their useful information may be represented correctly in feature space even when their pixelwise motion does not follow the assumptions required by a conventional optical-flow estimator. A temporal operator tied to dense image-space correspondence can consequently become misaligned with the source used in Stage~II. STAR-VF addresses this by decoding frame-compatible clips but establishing temporal dependency after encoding, as detailed in \appsecref{app:temporal}.

Compression further concentrates informative cues. Salient structure, texture, illumination, and modality-specific contrast need not be distributed uniformly among the retained tokens. A global average or uniform pooling step can weaken precisely the localized evidence that distinguishes the two modalities. The query-driven aggregation in \appsubsecref{app:selective_aggregation} lets multiple learned slots select complementary subsets of the enhanced center-frame representation instead of assigning equal importance to all elements.

These requirements are linked rather than independent. The compact source must expose ordered, modality-matched inputs; the temporal module must operate on a representation compatible with those inputs; and the aggregation module must prevent the retained evidence from being diluted before reconstruction. This chain motivates the canvas-decoder pair below and the feature-level objective in \appsecref{app:training}.

\subsection{Condensed Source and Canvas Decoder}
\label{app:canvas_decoder}

STAR-VF replaces the original training set with $N_s$ learnable paired samples $\mathcal{S}=\{(\mathbf{G}_{1,s},\mathbf{G}_{2,s})\}_{s=1}^{N_s}$. For modality $k$, each canvas is arranged as a $\rho\times\rho$ grid and decoded in row-major order:
\begin{equation}
\begin{aligned}
\mathbf{G}_{k,s}&\in\mathbb{R}^{C_k\times\rho P_h\times\rho P_w},
\qquad u=(a-1)\rho+b,\\
\widetilde{\mathcal{V}}_{k,s}
&=\mathcal{U}(\mathbf{G}_{k,s};k)
=\{\widetilde{\mathbf{I}}_{k,s,u}\}_{u=1}^{\rho^2},
\quad
\widetilde{\mathbf{I}}_{k,s,u}\in\mathbb{R}^{C_k\times H\times W}.
\end{aligned}
\label{eq:app_canvas_decoder}
\end{equation}
The modality argument in $\mathcal{U}(\cdot;k)$ allows one decoder design to produce outputs with the statistics and channel layout expected by each source modality. The grid contains $T_{\mathrm{cond}}=\rho^2$ ordered units, so the decoded output can be sampled as a short clip rather than as an unordered collection of images.

The pair $(\mathcal{S},\mathcal{U})$ specifies both what compact support is retained and how that support is realized as frame-shaped training data. In Stage~I, decoded frames can be passed through the same frozen proxy as real frames. In Stage~II, decoded clips can be grouped into windows of the form in \appeqref{eq:app_local_operator}. The decoder is therefore part of the training representation, not a visualization module: without it, the latent canvas would be incompatible with both the proxy feature extractor and the frame-based fusion network.

Let $N_{\mathrm{real}}$ be the number of paired temporal units in the original training set. The effective training-source ratio is $r_{\mathrm{cond}}=N_s\rho^2/N_{\mathrm{real}}$. This definition counts decoded temporal units rather than pixels. It separates data condensation from spatial downsampling: the decoded frames retain the input resolution required by the final model, while the number of learned training units is reduced. The experiments use $r_{\mathrm{cond}}=1/16$, as reported in \apptabref{tab:condensation_budget}.

Once Stage~II is complete, neither $\mathcal{S}$ nor $\mathcal{U}$ is needed for inference. The deployed system is $\mathcal{F}_{\Theta}$ operating on real paired videos. The condensed branch changes the cost and composition of training but does not add a decoding step to the test-time path.

\subsection{Temporal Ordering and Window Coverage}
\label{app:window_coverage}

The row-major mapping in \appeqref{eq:app_canvas_decoder} gives each grid cell a unique temporal index $u\in\{1,\ldots,\rho^2\}$. The two modalities use the same indexing rule, so $(\widetilde{\mathbf{I}}_{1,s,u},\widetilde{\mathbf{I}}_{2,s,u})$ forms a paired temporal unit. This shared order is required for fusion: modality complementarity is defined at a common index, and temporal neighborhoods are constructed by advancing that index consistently in both decoded clips. The spatial arrangement of cells inside $\mathbf{G}_{k,s}$ is a compact parameterization; after decoding, their row-major indices provide the sequence order used by Stage~II.

For a decoded clip of length $\rho^2$ and an odd local window length $L$, every center index can produce a window when the boundary rule in \appsubsecref{app:local_formulation} is applied. If only fully contained windows are retained, one clip provides $\rho^2-L+1$ center positions. Thus, $\rho$ affects not only the number of decoded frames but also the diversity of local temporal contexts available from each canvas. With $N_s$ canvases, the number of candidate centers scales linearly with $N_s$ and approximately quadratically with $\rho$, before any stochastic sampling used during training.

Ordering does not imply that adjacent grid cells must resemble neighboring image patches. The decoder maps each compact unit to a full-resolution frame, and temporal supervision is supplied by the ordered windows built from those outputs. This separates storage geometry from prediction geometry: the canvas uses a two-dimensional layout to share a compact decoder, whereas the fusion model operates on full-resolution frames and token coordinates. Consequently, the spatial coordinates $(x_j,y_j)$ in M-RoPE refer to positions within decoded frames, not positions of cells inside the canvas.

This distinction prevents two common ambiguities. First, $\rho\times\rho$ is not the resolution of a synthetic frame; every decoded frame has resolution $H\times W$. Second, $T_{\mathrm{cond}}=\rho^2$ is the temporal support per condensed sample, not the length of a real test video. The real sequence can contain any $T$ supported by the sliding inference procedure, while the compact source only needs to supply local training contexts of length $L$.

\input{Figure/supp_gridsize_tradeoff}

\subsection{Canvas Capacity and Grid Size}
\label{app:canvas_design}

A direct synthetic-video parameterization would learn $T_{\mathrm{cond}}$ full-resolution frames per modality. A frame-wise compact alternative would learn an independent $C_k\times P_h\times P_w$ tensor for every synthetic time step. STAR-VF stores the same compact support in one structured canvas. Their parameter counts are
\begin{equation}
N_{\mathrm{direct}}=C_kHW T_{\mathrm{cond}},
\qquad
N_{\mathrm{frame}}=C_kP_hP_w T_{\mathrm{cond}},
\qquad
N_{\mathrm{canvas}}=C_k\rho^2P_hP_w.
\label{eq:app_capacity_counts}
\end{equation}
With $T_{\mathrm{cond}}=\rho^2$, $N_{\mathrm{canvas}}=N_{\mathrm{frame}}$. The gain over independent compact units is therefore structural rather than a claim of fewer scalar variables: all units belong to one jointly optimized object and are interpreted by a shared decoder. Relative to direct full-resolution clips, the reduction factor per modality is $P_hP_w/(HW)$ for the same number of temporal units.

The grid size $\rho$ controls three quantities simultaneously: latent capacity, decoded clip length, and the number of windows available to Stage~II. Small grids impose stronger compression but provide fewer temporal configurations. Larger grids provide more synthetic support and allow the decoder to store more variation, but weaken compactness and can allocate capacity to redundant patterns. Consequently, increasing $\rho$ need not improve the final model monotonically.

The ablation in \apptabref{tab:ablation2} follows this trade-off, and \appfigref{fig:grid_size_tradeoff} shows its geometric interpretation. The $4\times4$ setting supplies enough ordered units for local temporal training while maintaining the target data budget. The selected value is therefore determined by downstream performance under compression, not by treating the canvas as an image whose spatial resolution should simply be maximized.

\newpage
\section{Flow-Free Temporal Modeling and Selective Aggregation}
\label{app:temporal}

The final model first establishes temporal dependency within each modality and then extracts a compact set of fusion-relevant elements from the enhanced center-frame representation. Both operations are defined in feature space, which keeps them compatible with decoded condensed clips and preserves a direct path from modality-specific evidence to reconstruction.

\subsection{Representation Compatibility and Flow Sensitivity}
\label{app:flow_sensitivity}

Temporal modeling under condensation can be implemented with flow-based warping, 3D convolution, recurrent or state-space propagation, or local attention. These choices differ mainly in the assumptions imposed on the input representation. Flow-based warping expects recoverable image-space correspondence. A 3D convolution uses fixed local kernels over space and time. Sequential propagation compresses previous observations into a state and introduces an asymmetric update order. Local attention instead relates the center frame to all encoded tokens in a bounded neighborhood and lets the relevance weights depend on their content and position.

Decoded clips are optimized to preserve fusion-relevant statistics, not dense optical-flow fields. If a decoded frame is written conceptually as $\widetilde{\mathbf{I}}_{k,\tau}=\mathbf{I}^{\star}_{k,\tau}+\boldsymbol{\varepsilon}_{k,\tau}$, where $\boldsymbol{\varepsilon}_{k,\tau}$ denotes decoding variation, a conventional alignment pipeline would use
\begin{equation}
\widehat{\boldsymbol{\Delta}}_{\tau\rightarrow t}
=\operatorname{Flow}(\widetilde{\mathbf{I}}_{k,\tau},\widetilde{\mathbf{I}}_{k,t}),
\qquad
\widetilde{\mathbf{X}}_{k,\tau\rightarrow t}
=\operatorname{Warp}(\mathbf{X}_{k,\tau},\widehat{\boldsymbol{\Delta}}_{\tau\rightarrow t}).
\label{eq:app_flow_pipeline}
\end{equation}
Appearance variation first affects the displacement estimate and is then propagated through the warp. Since the Stage~I objective in \mainequref{eq:condensation_loss} does not supervise dense correspondence, this introduces an interface whose assumptions are not enforced during condensation. It can also suppress useful modality-specific patterns if a locally plausible displacement is estimated from a synthetic appearance change rather than real motion.

STAR-VF removes that interface. Each frame is encoded independently, and temporal interaction is applied directly to the resulting tokens. The model can then use a neighboring token because it is relevant to the center representation, without first expressing that relation as a pixel displacement. This choice does not discard spatial organization: token position is included by M-RoPE in \appsubsecref{app:compact_attention}.

\subsection{Joint Spatio-Temporal Attention with M-RoPE}
\label{app:compact_attention}

For every $\tau\in\mathcal{N}(t)$, the modality-specific encoder produces $\mathbf{X}_{k,\tau}=\mathcal{E}_k(\mathbf{I}_{k,\tau})\in\mathbb{R}^{J\times d}$, where $J$ is the number of tokens. The center frame supplies the query, while all frames in the window supply keys and values. For head $m$, the projections are
\begin{equation}
\mathbf{Q}_{k,t}^{(m)}=\mathbf{X}_{k,t}\mathbf{W}_{Q}^{(m)},
\quad
\mathbf{K}_{k,\tau}^{(m)}=\mathbf{X}_{k,\tau}\mathbf{W}_{K}^{(m)},
\quad
\mathbf{V}_{k,\tau}^{(m)}=\mathbf{X}_{k,\tau}\mathbf{W}_{V}^{(m)}.
\label{eq:app_temporal_qkv}
\end{equation}
This center-query construction keeps the output aligned with the target frame. Neighboring frames act as temporal memory rather than as additional output positions.

For token $j$, M-RoPE uses the coordinate $\mathbf{p}_{\tau,j}=(\tau,x_j,y_j)$ and applies temporal and spatial rotations through $\Psi(\cdot,\mathbf{p}_{\tau,j})$. The encoded keys and unrotated values are concatenated across the $L$ frames into $\overline{\mathbf{K}}_{k,t}^{(m)},\overline{\mathbf{V}}_{k,t}^{(m)}\in\mathbb{R}^{LJ\times d_h}$. Attention is then
\begin{equation}
\begin{aligned}
\mathbf{A}_{k,t}^{(m)}&=\operatorname{Softmax}_{LJ}\!\left(
\frac{\Psi(\mathbf{Q}_{k,t}^{(m)},\mathbf{p}_{t})
\overline{\mathbf{K}}_{k,t}^{(m)\top}}{\sqrt{d_h}}
\right),\\
\mathbf{H}_{k,t}^{(m)}&=\mathbf{A}_{k,t}^{(m)}\overline{\mathbf{V}}_{k,t}^{(m)},
\qquad
\mathbf{Z}_{k,t}=\operatorname{Proj}\!\left[\operatorname{Concat}_{m=1}^{h}
\mathbf{H}_{k,t}^{(m)}\right].
\end{aligned}
\label{eq:app_mrope_attention}
\end{equation}
The softmax in \appeqref{eq:app_mrope_attention} is normalized over all $LJ$ memory tokens, not separately over frames. Tokens from different times therefore compete in one distribution, while the response scale remains independent of window length. This is the same computation formalized in \mainequrefrange{eq:temporal_memory}{eq:temporal_enhanced_representation}.

The position vector carries both temporal order and spatial location. Content-only attention could assign the same relation to visually similar tokens before and after the center time; a temporal-only position term would distinguish order but omit where a token occurs in the frame. M-RoPE retains both coordinates, so the similarity score can distinguish preceding from following evidence and preserve the spatial geometry required by fusion. The resulting $\mathbf{Z}_{k,t}\in\mathbb{R}^{J\times d}$ is an enhanced center-frame representation with information selected from the whole local window.

The tensor shapes also clarify what is and is not compressed by the temporal block. A query head contains $J$ center-frame tokens, while its key and value memories contain $LJ$ tokens. Multiplication by $\overline{\mathbf{K}}_{k,t}^{(m)\top}$ therefore produces a $J\times LJ$ score matrix. Each row describes how one center token draws evidence from every spatial position and time step in the local memory. Multiplication by $\overline{\mathbf{V}}_{k,t}^{(m)}$ returns $J\times d_h$, so the number and spatial indexing of center tokens are preserved through temporal enhancement.

Joint normalization over $LJ$ tokens is important for comparisons across frames. If attention were normalized separately inside each frame and the frame responses were summed, every frame would contribute a unit-normalized response regardless of its relevance, and the output magnitude would grow with $L$ unless another weighting stage were added. The normalization in \appeqref{eq:app_mrope_attention} instead allocates one probability mass across the entire memory. A nearby frame can dominate when it contains decisive evidence, while an uninformative frame can receive little weight.

M-RoPE enters only the query-key similarity. Values remain content features and are mixed according to the position-aware weights. This keeps the retrieved evidence in the learned feature basis while using $(\tau,x_j,y_j)$ to determine where it should be retrieved from. The final projection combines the heads and returns the original channel dimension $d$, making $\mathbf{Z}_{k,t}$ directly compatible with the aggregation projections in \appeqref{eq:app_query_aggregation}.

\subsection{Local Temporal Scope}
\label{app:local_scope}

The effect of restricting attention to $\mathcal{N}(t)$ can be expressed by comparing a full-sequence response with the omitted contribution:
\begin{equation}
\mathbf{H}_{k,t}^{(m),\mathrm{full}}
=\sum_{\tau=1}^{T}\mathbf{A}_{k,t\leftarrow\tau}^{(m)}\mathbf{V}_{k,\tau}^{(m)},
\quad
\mathbf{H}_{k,t}^{(m)}
=\sum_{\tau\in\mathcal{N}(t)}\mathbf{A}_{k,t\leftarrow\tau}^{(m)}\mathbf{V}_{k,\tau}^{(m)},
\quad
\mathbf{R}_{k,t}^{(m)}
=\sum_{\tau\notin\mathcal{N}(t)}\mathbf{A}_{k,t\leftarrow\tau}^{(m)}\mathbf{V}_{k,\tau}^{(m)}.
\label{eq:app_local_residual}
\end{equation}
The local approximation is accurate when $\mathbf{R}_{k,t}^{(m)}$ is small relative to the evidence inside the window. This condition is consistent with the evaluated tasks, where temporal quality is dominated by adjacent motion, local structural continuity, and short-range illumination change. It also matches the synthetic source, whose ordered support is intentionally compact.

The choice of $L$ balances information coverage against computation and relevance. A short window can miss useful motion or complementary evidence. A much longer window increases attention cost and introduces distant tokens whose relation to the center may be weak or ambiguous. Joint normalization can then distribute probability mass over redundant context. The selected local scope focuses the operator on the range most directly supervised by the decoded clips.

Because \appeqref{eq:app_local_operator} is evaluated at every $t$, local attention still produces a full-sequence result. Its limitation is temporal reach, not output length. Scenes with long occlusion, delayed cross-modal evidence, or large non-local displacement can yield a non-negligible residual; these cases are discussed in \appsubsecref{app:temporal_limitations}.

\subsection{Query-Driven Intra-Frame Aggregation}
\label{app:selective_aggregation}

Temporal attention enhances $J$ center-frame tokens, but those tokens do not contribute equally to fusion. Some carry boundaries or fine texture, others capture thermal saliency or exposure detail, and many describe regions already represented adequately by the other modality. Uniform pooling would compress these elements with fixed weights and can dilute localized evidence after the already aggressive reduction of the training source.

STAR-VF uses $M$ learned aggregation queries per modality. After linear projection, each query attends over all $J$ enhanced tokens:
\begin{equation}
\mathbf{Y}_{k,t}=\operatorname{Softmax}_{J}\!\left(
\frac{(\mathbf{Q}^{\mathrm{agg}}_k\mathbf{W}^{k}_{Q})
(\mathbf{Z}_{k,t}\mathbf{W}^{k}_{K})^{\top}}{\sqrt d}
\right)(\mathbf{Z}_{k,t}\mathbf{W}^{k}_{V})
\in\mathbb{R}^{M\times d}.
\label{eq:app_query_aggregation}
\end{equation}
The row-wise softmax in \appeqref{eq:app_query_aggregation} gives every slot its own selection pattern. One slot may emphasize high-contrast structure, another may collect smooth intensity evidence, and another may respond to localized modality-specific content. The roles are learned from the fusion objective rather than assigned manually. This produces a structured summary instead of collapsing all tokens into one global vector.

The two modality summaries retain separate parameters until aggregation is complete, so evidence is selected within its native representation before cross-modal reconstruction. They are then concatenated along the feature dimension and passed to $\mathcal{D}$, exactly as defined in \mainequrefrange{eq:qkv_aggregation}{eq:reconstruction}. This ordering supports complementarity: each modality contributes several selected elements, and the decoder reconstructs the fused center frame from their joint representation.

Using multiple slots also avoids forcing heterogeneous cues into a single bottleneck. A one-vector summary would require structure, texture, intensity, and modality-specific saliency to share the same attention pattern. With $M$ rows, $\mathbf{Y}_{k,t}$ can retain several evidence types while remaining substantially smaller than the $J$-token input when $M<J$. The output is therefore compact in token count but not uniform in content. The aggregation queries are model parameters learned in Stage~II; they are distinct from the condensed canvases optimized in Stage~I.

Cross-modal interaction occurs after each modality has completed temporal enhancement and selective summarization. This preserves the meaning of modality-specific tokens during selection and gives the reconstruction decoder matched sets of $M$ slots. Concatenation along the feature dimension pairs the summaries without averaging them, allowing $\mathcal{D}$ to combine corroborating evidence or retain a detail that appears strongly in only one source. The reconstruction objective in \mainequref{eq:fusion_loss} supplies the signal that determines which slot patterns are useful.

The temporal and intra-frame modules therefore solve different selection problems. \appeqref{eq:app_mrope_attention} determines which tokens across neighboring frames improve the center representation. \appeqref{eq:app_query_aggregation} determines which elements of that enhanced representation should survive the compact summary used for fusion. Keeping these stages distinct preserves temporal order without forcing the reconstruction module to process every token uniformly.

\newpage
\section{Condensation Objectives and Two-Stage Training}
\label{app:training}

Stage~I learns a compact source that preserves statistics useful to the downstream fusion model. Stage~II freezes that source and trains the final architecture with the same fusion objective used for real video inputs. This section details the exact matching terms and the boundary between the two stages.

\subsection{Feature-Space Supervision}
\label{app:feature_matching}

A direct pixel objective would require every decoded frame to reproduce a particular real frame even when several appearances provide equivalent training signals. Under a $1/16$ source budget, this would spend limited capacity on sample-specific pixel configurations rather than on statistics shared by the task. STAR-VF instead uses a frozen proxy $\phi_{k,\ell}$ to map real and synthetic frames to a common feature hierarchy:
\begin{equation}
\mathbf{F}^{r}_{k,\ell,n,t}=\phi_{k,\ell}(\mathbf{I}^{r}_{k,n,t}),
\qquad
\mathbf{F}^{s}_{k,\ell,s,u}=\phi_{k,\ell}(\widetilde{\mathbf{I}}_{k,s,u}).
\label{eq:app_proxy_features}
\end{equation}
The proxy remains fixed, so the representation against which the canvases and decoder are optimized does not drift. Matching multiple levels covers complementary properties: shallower activations retain edges, contrast, and modality-specific texture, while deeper activations capture region organization and salient structure. The condensed source is not required to match each real sample one-to-one; it matches pooled statistics across sampled clips and frames.

\subsection{Exact Multi-Level Matching Objective}
\label{app:condensation_loss}

The Stage~I loss is the same objective given in \mainequref{eq:condensation_loss}:
\begin{equation}
\mathcal{L}_{\mathrm{cond}}
=\underbrace{\mathcal{D}^{16}_{\mathrm{mse}}
+\lambda_{1}\mathcal{D}_{\mathrm{ch}}
+\lambda_{2}\!\left(\mathcal{D}^{4}_{\mathrm{sp}}+\mathcal{D}^{8}_{\mathrm{sp}}\right)}_{\text{multi-level feature matching}}
+\underbrace{\lambda_{3}\mathcal{L}_{\mathrm{color}}}_{\text{pixel-space matching}}.
\label{eq:app_condensation_loss}
\end{equation}
Here, $\mathcal{D}^{16}_{\mathrm{mse}}$ compares mean proxy feature maps after alignment to a $16\times16$ spatial support. It preserves average feature organization instead of reducing the representation immediately to a single global statistic. The channel term $\mathcal{D}_{\mathrm{ch}}$ compares the mean and standard deviation of every channel, retaining activation strength and dispersion. The spatial terms $\mathcal{D}^{q}_{\mathrm{sp}}$ compare mean and standard deviation within $q\times q$ grids for $q\in\{4,8\}$, preserving coarse and intermediate spatial layouts. All terms are accumulated over modalities and proxy levels after pooling the sampled real and synthetic units.

The three feature statistics are complementary. A global channel match can reproduce the distribution of responses while placing them in incorrect regions. A spatial mean can preserve layout but omit the variation carried by feature dispersion. The $16\times16$ proxy-map term retains finer organization, while the $4\times4$ and $8\times8$ statistics stabilize structure at broader scales. Their combination constrains what is present, how strongly it is represented, and where it occurs without requiring a one-to-one pairing between real and synthetic frames.

Operationally, the real side aggregates statistics over the sampled batch, frame indices, and spatial support appropriate to each term. The synthetic side applies the same reductions to decoded units. A mean statistic aligns the center of the response distribution, while a standard-deviation statistic aligns its spread. Computing both prevents a constant synthetic activation from matching a non-trivial real distribution through its mean alone. Applying the reductions at several spatial granularities similarly prevents a globally correct statistic from masking a regionally incorrect arrangement.

The objective is symmetric with respect to the two source modalities in its aggregation structure, but the proxy features and decoded frames remain modality-specific. This is important when the sources have different channel counts or appearance statistics: the loss does not force infrared and visible inputs, for example, to become identical. Instead, each synthetic modality is matched to the corresponding real modality before the two are combined by the fusion model. Pairing is preserved through the common indices $s$ and $u$ introduced in \appeqref{eq:app_canvas_decoder}.

Feature extractors are intentionally tolerant to some low-level changes. Without an image-space constraint, the decoder could match proxy statistics while drifting in brightness, contrast, or modality-specific channel distribution. $\mathcal{L}_{\mathrm{color}}$ applies the corresponding moment matching in pixel space and anchors the output of $\mathcal{U}$ to a stable appearance regime. This term is especially relevant at the transition to Stage~II, where the final encoders receive decoded images rather than proxy features. It prevents the final model from learning around avoidable artifacts of the condensed branch.

The weights $\lambda_1$, $\lambda_2$, and $\lambda_3$ balance different statistics; they do not introduce a separate target task. Together, the terms in \appeqref{eq:app_condensation_loss} preserve the spatial, channel-wise, and low-level properties needed by the fixed input interface defined in \appeqref{eq:app_canvas_decoder}.

\subsection{Two-Stage Optimization}
\label{app:two_stage}

Jointly updating the condensed source and the final model would make the training distribution and its consumer change simultaneously. The source could adapt to transient weaknesses of the current fusion network, while the network could adapt to transient decoder artifacts. STAR-VF separates the problems:
\begin{equation}
(\mathcal{S}^{*},\mathcal{U}^{*})
=\arg\min_{\mathcal{S},\mathcal{U}}\mathcal{L}_{\mathrm{cond}},
\qquad
\mathcal{F}_{\Theta}^{*}
=\arg\min_{\Theta}\mathcal{L}_{\mathrm{fuse}}(\Theta;\mathcal{S}^{*},\mathcal{U}^{*}).
\label{eq:app_two_stage}
\end{equation}
Stage~I determines which supervision survives compression. Its optimization variables are the canvases and the modality-conditioned decoder, while the proxy is fixed. After convergence, $(\mathcal{S}^{*},\mathcal{U}^{*})$ is frozen. Stage~II initializes the final fusion model from scratch and determines how the retained supervision is converted into fused outputs. This separation ensures that performance is attributable to a reusable training source rather than a private co-adaptation between one synthetic set and one intermediate model state.

Stage~II builds ordered local windows from $\widetilde{\mathcal{V}}_{k,s}$ and applies the full architecture in \mainalgref{alg:overall_framework}. The prediction loss remains the intensity, gradient, SSIM, and temporal combination in \mainequref{eq:fusion_loss}. The first three components preserve complementary content in each output frame; the temporal component regularizes changes across adjacent predictions. Condensation changes the training examples supplied to this objective, not the definition of successful fusion.

The gradient boundary between the stages is explicit. During Stage~I, gradients pass from the matching statistics through the frozen proxy to $\mathcal{U}$ and $\mathcal{S}$, but proxy parameters are not updated. During Stage~II, decoded clips are treated as fixed data: gradients update the encoders, temporal attention, aggregation queries, and reconstruction decoder inside $\mathcal{F}_{\Theta}$, but do not modify $\mathcal{S}^{*}$ or $\mathcal{U}^{*}$. At inference, neither set of Stage~I variables participates in the forward pass.

This schedule also makes evaluation reproducible across model initializations. A single learned source can be frozen and used to train a fresh fusion model under the same protocol as alternative compact sources. The random-source comparison in \appsubsecref{app:random_results} follows this separation; differences in the final metrics can therefore be attributed to the training source rather than to simultaneous adaptation of that source during Stage~II.

\subsection{Compatibility across Training and Inference}
\label{app:training_compatibility}

The learned source is compatible with the final model at three levels. First, interface compatibility follows from \appeqref{eq:app_canvas_decoder}: decoded frames have the modality-specific channels and spatial resolution of real inputs, and their ordering supports the same local-window construction. No synthetic-only encoder or test-time adapter is introduced.

Second, representation compatibility follows from \appeqref{eq:app_condensation_loss}. The proxy-map, channel, spatial, and color terms preserve the statistics encountered by the early and intermediate stages of the final model. The synthetic frames can differ from particular real samples while remaining inside a useful training regime. Query-driven aggregation then selects non-uniform evidence from each modality without assuming that synthetic content is evenly distributed.

Third, objective compatibility follows from keeping $\mathcal{L}_{\mathrm{fuse}}$ unchanged. The model trained on condensed windows is evaluated on real windows with the same input-output relation in \appeqref{eq:app_local_operator}. At inference, the frozen condensed branch is removed; the real paired sequence is encoded, temporally enhanced, selectively aggregated, and reconstructed by $\mathcal{F}_{\Theta}^{*}$. The efficiency gain is therefore a reduction in training data, while the deployed architecture remains the fusion network described in \mainsecref{sec:method}.

The transition between stages is summarized as a stable sequence: learn a task-aligned source against a fixed proxy, freeze that source, train a fresh model against a fixed fusion objective, and retain only the model. This boundary also clarifies the limitation in \appsubsecref{app:representation_limitations}: Stage~II can exploit the supervision preserved by Stage~I but cannot recreate information absent from the condensed source.

\newpage
\section{Additional Experimental Results}
\label{app:additional_results}
\input{section/supp_random_box}
The reported metrics describe complementary aspects of video fusion. EN, CC, and SSIM primarily measure information content, agreement, and structural quality at the frame level, with larger values preferred. BiSWE and MS2R characterize temporal behavior, with smaller values preferred in the reported protocol. A temporal score should therefore be interpreted together with spatial quality: an output can vary little because it is stable, or because useful detail has been smoothed away. The comparisons below use the full metric profile to distinguish those cases.

\subsection{Random Initialization \textit{vs.} Learned Condensed Data}
\label{app:random_results}

\input{table/supp_random}

\apptabref{tab:random_condensation} compares five randomly constructed compact sources with the learned condensed source under the same data budget and Stage~II protocol. Across the random runs, EN, CC, and SSIM have means of $5.64$, $0.48$, and $0.51$, with standard deviations of $0.79$, $0.12$, and $0.08$. The learned source reaches $6.73$, $0.64$, and $0.65$, respectively. It therefore lies near the strong end of the random distribution on all three image-level metrics rather than benefiting on only one measure.

The individual runs also show why a fixed compact budget is insufficient by itself. Random\_2 obtains EN $4.76$, CC $0.43$, and SSIM $0.46$, whereas Random\_3 reaches $6.80$, $0.67$, and $0.65$. Such dispersion means that the quality of the retained support depends strongly on the sampled initialization when no condensation objective organizes it. Learned canvases reduce this uncertainty by optimizing the spatial, channel, and color statistics that are later consumed by the final model.

\input{section/supp_random_std}

\appfigref{fig:random_boxplot} shows the individual-run distribution after direction-aligned normalization, while \appfigref{fig:random_errorbar} summarizes its mean and variance. The visualizations place the learned source relative to the distribution rather than relative to a single selected seed. This matters because an isolated favorable run can conceal the instability of random compact construction.

The temporal metrics should be read jointly with the spatial metrics. Some random runs obtain lower BiSWE or MS2R, but these cases coincide with markedly weaker EN, CC, or SSIM. A low temporal-variation score can result from outputs that change little because informative detail has been suppressed. The learned source instead provides a balanced point: it preserves substantially stronger spatial content while keeping temporal behavior within the range of the compact baselines. The comparison therefore isolates the effect of optimized condensation from the effect of training on a small source alone.

\subsection{Additional Baselines on VTMOT}
\label{app:additional_baselines}

\input{table/supp_img}

\apptabref{tab:vtmot_only} extends the VTMOT comparison with representative image-fusion methods. GIFNet obtains the highest CC, while SwinFusion obtains the highest EN and SSIM. These results confirm that single-frame models can retain strong per-frame complementary information on this dataset. Their image-level strengths provide an informative reference for evaluating whether explicit temporal modeling sacrifices spatial quality.

The temporal columns reveal a different pattern. GIFNet and TC-MoA obtain BiSWE values of $5.52$ and $4.69$, respectively, even though their frame-wise metrics are competitive. CDDFuse and SwinFusion also do not dominate the spatial and temporal measures simultaneously. Processing every frame independently leaves temporal regularity to emerge indirectly from similar inputs; it does not explicitly connect the prediction at $t$ to neighboring evidence.

STAR-VF reaches CC $0.64$, EN $6.73$, and SSIM $0.65$ together with BiSWE $3.91$ and MS2R $0.35$, using $1/16$ of the original training data. Its purpose is not to maximize one column in isolation but to retain a competitive spatial-temporal profile under a condensed source. Read together with \apptabref{tab:random_condensation}, the result supports two distinct conclusions: learned condensation is more reliable than random compact construction, and flow-free temporal modeling preserves video-level regularity that strong image baselines do not enforce directly.

The two comparisons answer different controls. Random initialization holds the architecture and compact budget fixed, isolating whether the learned source contains organized supervision. Image-fusion baselines change the architecture and use frame-wise processing, testing whether strong per-frame methods already provide the temporal balance observed for STAR-VF. The learned source improves the first control, while explicit temporal dependency addresses the second. Neither table alone establishes both effects; their joint reading connects the condensed representation to the temporal model without conflating them.

Metric differences near the top of a column are also best read as trade-offs rather than as a single total ordering. SwinFusion has slightly higher EN and SSIM, and GIFNet has higher CC, whereas their temporal profiles differ from STAR-VF. The result claimed here is therefore competitive preservation under substantial data reduction, supported by simultaneous spatial and temporal measurements. This interpretation is consistent with the capacity analysis in \appsubsecref{app:canvas_design}: the condensed representation is selected to preserve a balanced downstream signal rather than to reproduce every property of the full training set.

\newpage
\section{Limitations and Future Directions}
\label{app:limitations}

\subsection{Dependence on Condensed Representation Quality}
\label{app:representation_limitations}

The final model is trained only on frames produced by the learned pair $(\mathcal{S}^{*},\mathcal{U}^{*})$. Information omitted in Stage~I cannot be recovered by Stage~II, so the quality of the condensed representation sets an upper bound on downstream performance. A fixed source budget may underrepresent rare structures, unusual modality interactions, or motion patterns that occupy little probability mass in the original data. The proxy statistics in \appeqref{eq:app_condensation_loss} preserve broad properties of the training distribution, but they do not guarantee that every low-frequency event receives an individual synthetic counterpart.

This dependence becomes more important as the original distribution grows in diversity while $N_s$ and $\rho$ remain fixed. The decoder must then express more modes with the same number of learned units. Increasing the grid or the number of canvases can raise capacity, but it also changes $r_{\mathrm{cond}}$ and weakens the intended compression. An adaptive budget that allocates support according to motion, modality discrepancy, or sample rarity is a direct extension of the current fixed-ratio design.

Proxy choice also affects which equivalences are encouraged. A proxy that is insensitive to a fusion-critical pattern may allow the synthetic source to omit it even if the matching objective is minimized accurately. The multi-level and color terms reduce dependence on one representation depth, yet the condensed source remains task-aligned through the statistics selected for Stage~I. Evaluating alternative frozen proxies and cross-dataset transfer would clarify how broadly one learned source can support different fusion architectures.

These limitations suggest concrete diagnostic signatures. Loss of rare spatial content would appear as locally weakened boundaries or modality-specific details even when average metrics remain strong. Insufficient variation in the condensed source would appear as high sensitivity to scene type or as a gap between common and uncommon motion patterns. A proxy mismatch would appear when Stage~I statistics are well matched but Stage~II performance remains weak across independent model initializations. Separating these cases requires inspecting both decoded samples and downstream outputs rather than relying only on the final mean score.

The two-stage structure helps localize such failures. If multiple final models trained from the same source fail similarly, the bottleneck is likely the retained supervision or decoder. If performance varies widely across Stage~II initializations while the source is fixed, optimization of the final model becomes the more likely cause. This decomposition does not eliminate the representation ceiling, but it provides a clear experimental boundary for improving the appropriate stage.

\subsection{Temporal Scope and Non-Local Motion}
\label{app:temporal_limitations}

The residual in \appeqref{eq:app_local_residual} becomes significant when useful evidence lies outside $\mathcal{N}(t)$. Long occlusion, abrupt exposure change, large displacement, delayed modality complementarity, and temporally sparse events can all produce this condition. The output may remain locally smooth while missing a structure visible only in a more distant frame. This is a consequence of the bounded receptive field rather than of the absence of temporal modeling.

Highly non-local motion can also challenge token relevance within a short window. M-RoPE preserves temporal order and spatial coordinates, but it does not enlarge the observed interval. If an object crosses a large part of the frame or reappears after several missing observations, the available tokens may not contain a reliable chain of local evidence. Increasing $L$ is one solution, although the quadratic attention cost and the number of weakly related memory tokens also grow.

Hierarchical windows, sparse long-range memory, or adaptive neighborhood selection could extend temporal reach while preserving the local operator as the first stage. Such designs would need a corresponding condensed source with enough ordered support to train the added pathway. The representation and temporal backbone should therefore be expanded together rather than increasing sequence length in only one branch.

\subsection{Future Extensions}
\label{app:future_extensions}

Three directions follow from the preceding limitations. First, budget-aware condensation can vary $N_s$ or $\rho$ according to the diversity of a dataset or sample group. A simple extension could allocate more units to clusters with large motion or strong cross-modal discrepancy while preserving a global ratio. A more general formulation could learn the allocation jointly with the canvases under an explicit storage constraint.

Second, Stage~I can incorporate additional task-aligned statistics. Temporal feature moments, cross-modal correlation, or a lightweight sequence-level proxy could supplement the current spatial and color terms when the target domain contains broader dynamics. These additions should preserve the stable two-stage boundary: the proxy and matching criteria define the source, and the final fusion model is still trained afterward from a fixed initialization.

Third, the final model can adopt hierarchical attention, recurrent memory, or stronger video backbones to cover longer dependencies. The query-driven aggregation remains applicable because it operates on the enhanced center representation rather than on a specific temporal architecture. Together, these directions extend the amount and range of supervision represented by each stage while retaining the central principle of STAR-VF: learn a compact, frame-compatible supervisory source first, then train the deployed fusion model on that fixed source.

A stronger temporal backbone also changes what the condensed source must preserve. A model with long-range memory can use evidence that a short-window model never observes, but only if Stage~I supplies ordered variation from which that dependency can be learned. Expanding the backbone without expanding the temporal diversity of $\mathcal{S}$ can therefore leave the additional capacity under-supervised. Conversely, increasing $\rho$ without a model able to use the resulting contexts can spend the budget on redundant decoded units. Future scaling should coordinate the source budget, window construction, and temporal receptive field.

\subsection{Transfer and Scaling Considerations}
\label{app:transfer_scaling}

The current source is learned for a fixed task, modality pair, and proxy. Transfer to another fusion backbone is most plausible when the new model consumes the same frame interface and benefits from the same spatial and channel statistics. The two-stage design makes this question measurable: keep $(\mathcal{S}^{*},\mathcal{U}^{*})$ fixed, train several architectures from independent initializations, and compare their performance with models trained on the corresponding real data. A small and consistent gap would indicate that the condensed source captures architecture-independent supervision; a gap that grows for one backbone would indicate that its learning requirements are not represented adequately by the current matching objective.

Cross-dataset transfer is stricter because both appearance statistics and temporal patterns can change. Direct reuse is suitable when the target domain shares the modality semantics, resolution interface, and main motion regime. Larger shifts may require adapting the canvases, the decoder, or both. Initializing from an existing condensed source and continuing Stage~I on a target-domain sample is one practical extension: it preserves the compact parameterization while allowing proxy and color statistics to move toward the new distribution. The amount of adaptation required can be measured against training a target source from scratch under the same $r_{\mathrm{cond}}$.

Budget comparisons should report temporal units and model performance together. Because $r_{\mathrm{cond}}$ counts $N_s\rho^2$, two settings can have the same nominal ratio but organize it differently: more canvases provide greater sample diversity, while larger grids provide longer within-sample support. These allocations are not equivalent for a temporal model. Holding the ratio fixed while varying $(N_s,\rho)$ would separate diversity from sequence length and provide a more complete scaling analysis than changing $\rho$ alone.

The same principle applies beyond the evaluated modality pairs. A new source type can be incorporated when $\mathcal{U}(\cdot;k)$ produces the required channel format and the proxy supplies meaningful features for that modality. The condensation statistics should then be validated for the cues that drive the new fusion task rather than copied unchanged. This preserves the framework-level separation between compact source learning and final-model training while allowing the representation, proxy, and budget allocation to be specialized to the target domain.

\subsection{Practical Budget Selection}
\label{app:budget_selection}

A practical deployment can select the condensed budget by treating downstream quality and source size as a validation curve. Begin from a small $(N_s,\rho)$ pair that can form the required windows, learn the source with a fixed Stage~I protocol, and train the same Stage~II architecture for every candidate. The preferred point is the smallest ratio whose spatial-temporal profile remains within the desired tolerance of the full-data model. This criterion directly measures the purpose of condensation and avoids choosing a grid solely from reconstruction appearance.

When additional budget is available, increasing $N_s$ and increasing $\rho$ should be tested separately. More canvases can represent distinct scenes or modality relationships, whereas a larger grid lengthens the ordered support inside each sample. Datasets dominated by diverse scene content may benefit more from $N_s$; datasets dominated by longer motion patterns may benefit more from $\rho$. A controlled sweep at fixed $N_s\rho^2$ can reveal which form of capacity is limiting before the total ratio is increased.

Validation should preserve the two-stage boundary used for the final experiment. Every source candidate should be frozen before Stage~II, and the final model should be initialized independently. Using several Stage~II seeds distinguishes a source that consistently retains supervision from a favorable model initialization. Reporting the mean together with variation is particularly useful near the budget knee, where small sources may produce similar average scores but different training stability.

Finally, budget selection should include both per-frame and temporal measures. A setting that preserves EN, CC, or SSIM but degrades BiSWE or MS2R may lack ordered support; a setting with smooth temporal scores but weaker spatial measures may be over-compressed. Applying the joint interpretation introduced in \appsecref{app:additional_results} yields a compact source that supports the complete video-fusion objective rather than one metric family alone.

%% file: Figure/supp_gridsize_tradeoff.tex
\begin{figure}[b]
    \centering
    \includegraphics[width=0.63\linewidth]{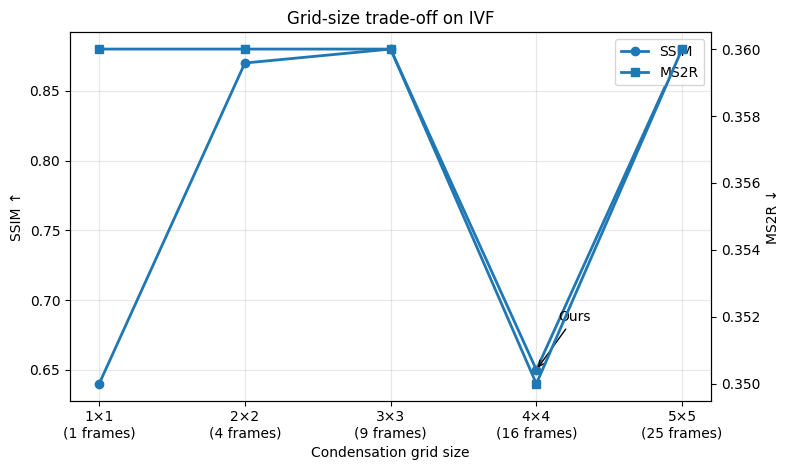}
    \caption{Illustration of the capacity--compactness trade-off controlled by the grid size $\rho$. Increasing $\rho$ enlarges both the latent canvas size and the decoded synthetic clip length $T_{\mathrm{cond}}=\rho^2$, thereby increasing the effective supervision budget available to Stage~II. At the same time, a larger $\rho$ weakens the compression effect by making the condensed source more flexible. The default setting $4\times4$ in STAR-VF is selected as a balanced operating point under this trade-off.}
    \label{fig:grid_size_tradeoff}
\end{figure}

%% file: section/supp_random_box.tex
\begin{figure}[b]
    \centering
    \includegraphics[width=0.63\linewidth]{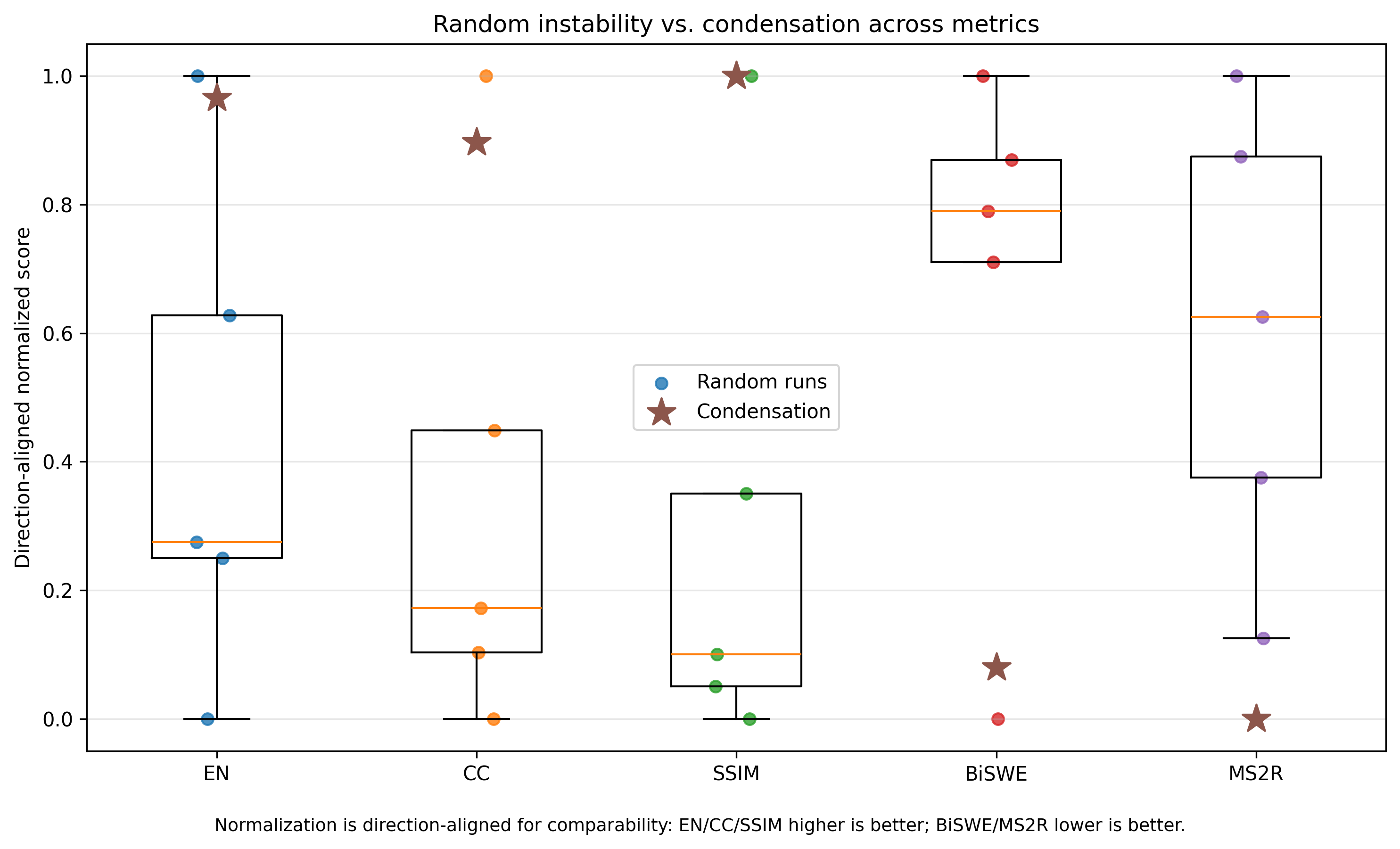}
    \caption{Distribution of random initialization across five metrics, with the condensation result overlaid as a reference point. For comparability across metrics with different scales and optimization directions, all values are transformed into direction-aligned normalized scores. A wider spread of the random runs indicates stronger seed sensitivity and lower stability.}
    \label{fig:random_boxplot}
\end{figure}

%% file: table/supp_random.tex
\begin{table*}[ht]
\centering
% \small
\setlength{\tabcolsep}{8pt}
\renewcommand{\arraystretch}{1.15}
\caption{Comparison between random initialization and condensation on VTMOT. Higher is better for EN, CC, and SSIM, while lower is better for BiSWE and MS2R.}
\label{tab:random_condensation}
\begin{tabular}{lccccc}
\toprule[1.2pt]
\textbf{Model} & \textbf{EN}$\uparrow$ & \textbf{CC}$\uparrow$ & \textbf{SSIM}$\uparrow$ & \textbf{BiSWE}$\downarrow$ & \textbf{MS2R}$\downarrow$ \\
\midrule
Random\_1    & 5.27 & 0.38 & 0.45 & 2.65 & 0.28 \\
Random\_2    & 4.76 & 0.43 & 0.46 & 2.83 & 0.29 \\
Random\_3    & 6.80 & 0.67 & 0.65 & 4.03 & 0.35 \\
Random\_4    & 5.32 & 0.41 & 0.47 & 2.94 & 0.31 \\
Random\_5    & 6.04 & 0.51 & 0.52 & 3.05 & 0.33 \\
\midrule
Random Mean  & 5.64 & 0.48 & 0.51 & 3.10 & 0.31 \\
Random Std   & 0.79 & 0.12 & 0.08 & 0.54 & 0.03 \\
\midrule
Condensation & 6.73 & 0.64 & 0.65 & 3.92 & 0.36 \\
\bottomrule[1.2pt]
\end{tabular}
\end{table*}

%% file: section/supp_random_std.tex
\begin{figure}[h]
    \centering
    \includegraphics[width=0.7\linewidth]{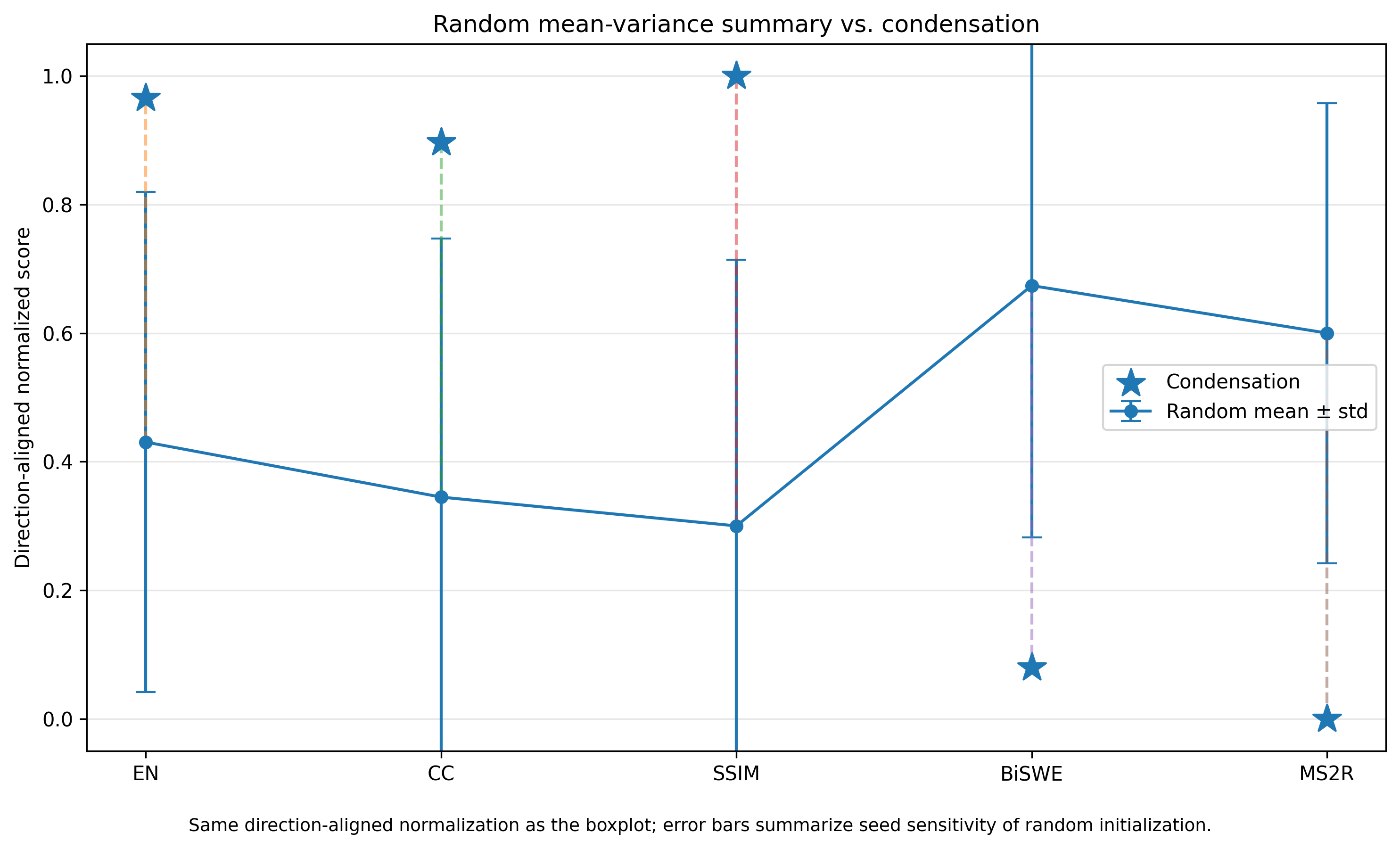}
    \caption{Mean-and-variance summary of random initialization across different metrics. Error bars denote the standard deviation over random runs, while the star marks the condensation result. The comparison shows that random initialization exhibits larger performance variability, whereas condensation provides a more stable and consistently strong outcome.}
    \label{fig:random_errorbar}
\end{figure}

%% file: table/supp_img.tex
\begin{table}[b]
\centering
% \footnotesize
% \scriptsize
\caption{Quantitative comparison on IVF (VTMOT~\cite{zhu2025visible}). 
\colorbox{red!20}{\textbf{Red}} and \colorbox{blue!15}{Blue} indicate the best and second-best results. Notably, STAR-VF is trained using only \textit{1/16} of the original training data, yet still achieves competitive performance on VTMOT.}
\vspace{-1.0em}
\label{tab:vtmot_only}
\setlength{\tabcolsep}{17pt}
\renewcommand{\arraystretch}{1.15}
\begin{tabular}{lccccc}
\toprule[1.2pt]
\textbf{Method}
& CC$\uparrow$ & EN$\uparrow$ & SSIM$\uparrow$ & BiSWE$\downarrow$ & MS2R$\downarrow$ \\
\midrule
GIFNet
& \cellcolor{red!20}0.67 & 6.41 & 0.56 & 5.52 & 0.38 \\
CDDFuse
& \cellcolor{blue!15}0.66 & \cellcolor{blue!15}6.77 & \cellcolor{blue!15}0.65 & 3.98 & \cellcolor{red!20}0.35 \\
TC-MoA
& 0.64 & 6.74 & \cellcolor{blue!15}0.65 & 4.69 & 0.38 \\
SwinFusion
& 0.65 & \cellcolor{red!20}6.81 & \cellcolor{red!20}0.66 & 3.93 & 0.37 \\
\midrule
UniVF~\citep{zhao2025unified}
& 0.65 & \cellcolor{red!20}6.81 & \cellcolor{blue!15}0.65 & 3.95 & \cellcolor{red!20}0.35 \\
VideoFusion~\citep{tang2025videofusion}
& 0.62 & 6.59 & 0.62 & \cellcolor{red!20}3.14 & 0.59 \\
TemCoCo~\citep{gong2025temcoco}
& 0.63 & 6.68 & 0.60 & 6.26 & 0.59 \\
MambaVF~\cite{zhao2026mambavf}
& - & - & \cellcolor{blue!15}0.65 & \cellcolor{blue!15}3.91 & \cellcolor{blue!15}0.36 \\
STAR-VF (ours)
& 0.64 & 6.73 & \cellcolor{blue!15}0.65 & \cellcolor{blue!15}3.91 & \cellcolor{red!20}0.35 \\
\bottomrule[1.2pt]
\end{tabular}
\end{table}